\documentclass{article}

\usepackage{microtype}
\usepackage{graphicx}
\usepackage{subcaption}
\usepackage{booktabs} 

\usepackage{hyperref}

\usepackage[preprint]{icml2026}

\usepackage{amsmath}
\usepackage{amssymb}
\usepackage{mathtools}
\usepackage{amsthm}
\usepackage{enumitem}
\usepackage{pifont}
\usepackage{dsfont}
\newcommand{\xmark}{\ding{55}}%
\usepackage{multirow}

\usepackage[capitalize,noabbrev]{cleveref}

\theoremstyle{plain}
\newtheorem{theorem}{Theorem}[section]

\theoremstyle{definition}

\theoremstyle{remark}

\newtheorem{property}{Property}
\usepackage[textsize=tiny]{todonotes}

\icmltitlerunning{\texttt{SVIP}: Towards Verifiable Inference of Open-source Large Language Models}

\begin{document}

\twocolumn[
  \icmltitle{\texttt{SVIP}: Towards Verifiable Inference of Open-source Large Language Models}



  \icmlsetsymbol{equal}{*}

  \begin{icmlauthorlist}
    \icmlauthor{Yifan Sun}{equal,yyy}
    \icmlauthor{Yuhang Li}{equal,yyy}
    \icmlauthor{Yue Zhang}{comp}
    \icmlauthor{Yuchen Jin}{comp}
    \icmlauthor{Huan Zhang}{yyy}
  \end{icmlauthorlist}

  \icmlaffiliation{yyy}{University of Illinois Urbana-Champaign}
  \icmlaffiliation{comp}{Hyperbolic Labs}

  \icmlcorrespondingauthor{Yifan Sun}{yifan50@illinois.edu}
  \icmlcorrespondingauthor{Huan Zhang}{huan@huan-zhang.com}

  \icmlkeywords{Machine Learning, ICML}

  \vskip 0.3in
]



\printAffiliationsAndNotice{\icmlEqualContribution}

\begin{abstract}
 The ever-increasing size of open-source Large Language Models (LLMs) renders local deployment impractical for individual users. \textit{Decentralized computing} has emerged as a cost-effective solution, allowing individuals and small companies to perform LLM inference for users using surplus computational power. However, a computing provider may stealthily substitute the requested LLM with a smaller, less capable model without consent from users, thereby benefiting from cost savings. We introduce \texttt{SVIP}, a secret-based verifiable LLM inference protocol. Unlike existing solutions based on cryptographic or game-theoretic techniques, our method is computationally effective and does not rest on strong assumptions. Our protocol requires the computing provider to return both the generated text and processed hidden representations from LLMs. We then train a proxy task on these representations, effectively transforming them into a unique model identifier. With our protocol, users can reliably verify whether the computing provider is acting honestly. A carefully integrated secret mechanism further strengthens its security. We thoroughly analyze our protocol under multiple strong and adaptive adversarial scenarios. Our extensive experiments demonstrate that \texttt{SVIP} is accurate, generalizable, computationally efficient, and resistant to various attacks. Notably, \texttt{SVIP} achieves false negative rates below  $5\%$ and false positive rates below  $3\%$, while requiring less than $0.01$ seconds per prompt query for verification.
\end{abstract}

\section{Introduction}
\label{sec:intro}
In recent years, open-source Large Language Models (LLMs) have achieved unprecedented success across a broad array of tasks and domains \citep{touvron2023llama, black2022gpt, le2023bloom, jiang2023mistral}, 
while remaining freely accessible. However, as model sizes increase, so do their computational demands \citep{kukreja2024literature}. As a result, \textbf{decentralized computing} \citep{uriarte2018blockchain} has gained significant attention as a cost-effective solution for users with limited local computational resources. In this setting, a user lacking computational power relies on \textit{decentralized} computing providers to perform LLM inference. These providers, often individuals or small companies with surplus resources, offer computational power at competitive prices. Commercial platforms facilitate such interactions by connecting both parties. Real-world examples include \href{https://www.golem.network/}{Golem Network}, 
\href{https://akash.network/}{Akash Network}, 
\href{https://rendernetwork.com/}{Render Network},
\href{https://www.spheron.network/}{Spheron Network},
\href{https://hyperbolic.xyz/}{Hyperbolic}, and
\href{https://vast.ai/}{Vast.ai}.


However, unlike reputable companies with well-established credibility, computation outputs from decentralized computing providers may not always be trustworthy. Specifically, to ease the deployment of LLM inference, computing providers often provide API-only access to users, hiding implementation details. A new risk arises in this setting: \emph{how to ensure that the outputs from a computing provider are indeed generated by the requested LLM}? For instance, a user might request the \texttt{Llama-3.1-70B} model for complex tasks, but a dishonest computing provider could substitute the smaller \texttt{Llama-2-7B} model for cost savings, while still charging for the larger model. The smaller model demands significantly less memory and processing power, giving the computing provider a strong incentive to cheat. 
Restricted by the black-box API access, it is difficult for the user to detect such substitutions. 

Without assurance that users are receiving the service they specified and paid for, they may lose trust and abandon the platform. To prevent this outcome and maintain profitability, the platform must ensure that user-specified models are faithfully executed.  This highlights the need for \textbf{verifiable inference}, a mechanism designed to ensure that the model specified by the user is the one used during inference. 

\vspace{-0.4cm}

\begin{figure*}[t!]
    \setlength{\abovecaptionskip}{0.5cm}
    \setlength{\belowcaptionskip}{-0.5cm}
    \centering
    \includegraphics[width=1.0\linewidth]{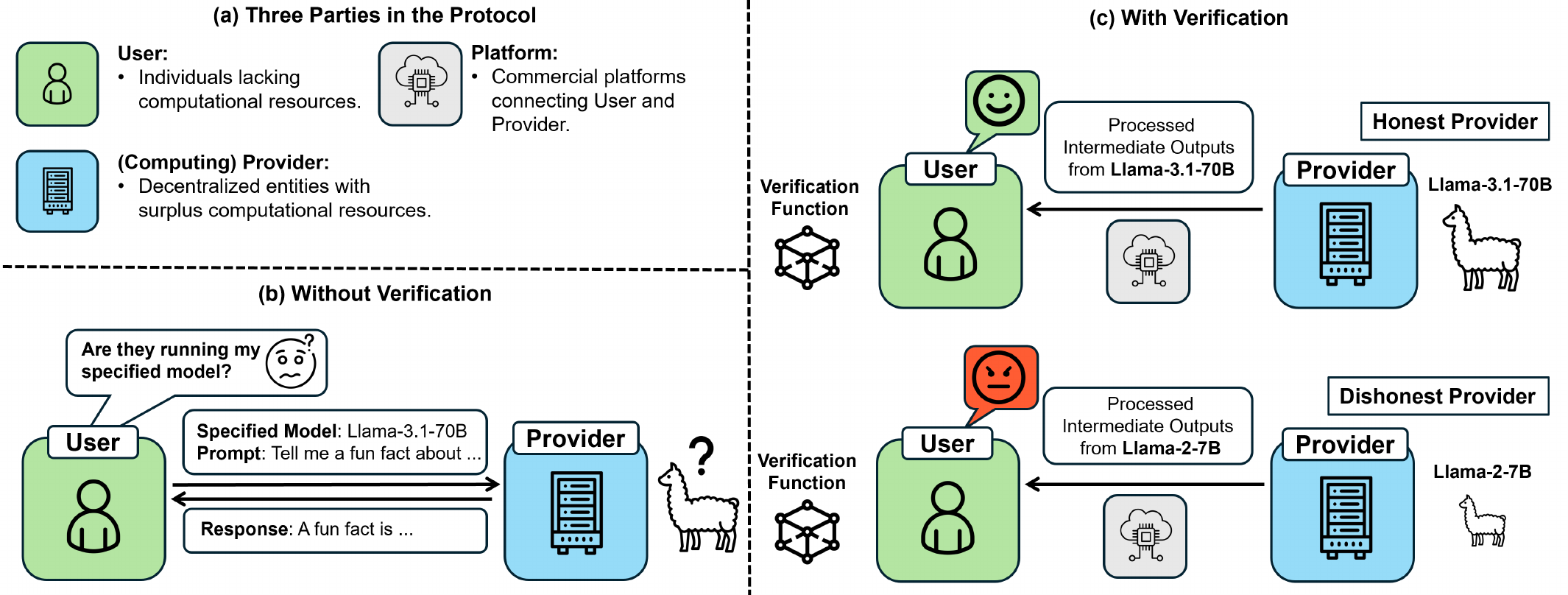}
    \caption{The problem setting of verifiable inference for LLMs. (a) Our protocol involves three parties. (b) A user requests the computing provider (referred to as \textit{provider} in the figure) to run inference on their prompt using the \texttt{Llama-3.1-70B} model. Without verification, they have no way to confirm if the specified model is used. (c) Our proposed protocol solves this by requiring the provider to return processed hidden representations from the LLM, enabling the user to verify through a verification function whether the correct model was used for inference. Specifically, the hidden representations are compressed to reduce the computational overhead.
    }
    \label{fig:problem_setting}
\end{figure*}


\paragraph{Related Work} A practical verifiable inference solution for LLMs must accurately confirm that the specified model is being used during inference while maintaining computational efficiency. Existing approaches face significant challenges that limit their applicability. Cryptographic verifiable computing methods, which rely on generating mathematical proofs \citep{yu2017survey, setty2012taking} or secure computation techniques \citep{gennaro2010non, laud2014verifiable} often impose high computational costs, making them unsuitable for real-time LLM inference. For instance, zkLLM, a recent Zero Knowledge Proof-based technique, requires over $803$ seconds for a single prompt query \citep{sun2024zkllm}. 
Game-theoretic protocols involve the interaction of multiple computing providers with carefully designed penalties and rewards \citep{zhang2024proof}, assuming all providers are rational, flawless, and non-cooperative, which might be unrealistic for certain system setups in practice. Meanwhile, watermarking and fingerprinting techniques \citep{kirchenbauer2023watermark, xu2024instructional} are mostly implemented by model publishers, making them unsuitable for verifiable inference, where the verification primarily occurs between the user and the computing provider. We leave an extended discussion of related work to Appendix \ref{sec:app_related}.


In this paper, we propose \texttt{SVIP}, a \textbf{\underline{S}}ecret-based \textbf{\underline{V}}erifiable LLM \textbf{\underline{I}}nference \textbf{\underline{P}}rotocol using hidden representations. Our protocol requires the computing provider to return not only the generated text but also the processed \textbf{hidden state representations} from the LLM. We carefully design and train a proxy task exclusively on the hidden representations produced by the specified model, effectively transforming these representations into a distinct identifier for that model. During deployment, users can verify whether the processed hidden states returned by the computing provider come from the specified model by assessing their performance on the proxy task. If the returned representations perform well on this task, it provides strong evidence that the correct model was used for inference. 
\textbf{Our key contributions are:}





\begin{itemize}[wide, labelwidth=!, noitemsep, labelindent=0pt,topsep=-\parskip]
    \item We systematically formalize the problem of verifiable LLM inference (\textsection\ref{sec: problem_statement}) and propose an innovative protocol that leverages processed hidden representations (\textsection\ref{sec: simple_protocol}). 
    \item The security of our protocol is further enhanced by a novel secret-based mechanism (\textsection\ref{sec:methods_secret}). We provide a thorough discussion and analysis of various strong and adaptive attack scenarios (\textsection\ref{sec:attacks}). 
    \item Our comprehensive experiments with $5$ specified open-source LLMs (from $13$B to $70$B) demonstrate the effectiveness of \texttt{SVIP}: it achieves an average false negative rate of $3.49 \%$, while keeping the false positive rate below $3 \%$ across $6$ smaller alternative models (\textsection\ref{sec:results_accuracy}). \texttt{SVIP} introduces negligible overhead (less than $0.01$ seconds per prompt query) for both users and computing providers (\textsection\ref{sec:cost_analysis}).  Furthermore, \texttt{SVIP} can effectively and securely handle $80$ to $120$ million prompt queries in total after a single round of protocol training, with the update mechanism further bolstering security (\textsection\ref{sec:results_attack}). 

\end{itemize}

\begin{table*}[t]
\centering
\setlength{\tabcolsep}{2pt}
\renewcommand{\arraystretch}{1.2}
\caption{Comparison of simple approaches and our proposed protocol based on the five criteria. A checkmark (\checkmark) indicates that the criterion is satisfied, while a cross (\xmark) indicates it is not. \texttt{SVIP} is the only method that satisfies all five criteria.}
\label{tab:naive_approaches}
\resizebox{0.8\linewidth}{!}{
\begin{tabular}{lccccc}
\toprule
\textbf{Approach} & \textbf{Low FNR} & \textbf{Low FPR} & \textbf{Efficiency} & \textbf{Completion Quality} & \textbf{Robustness} \\
\midrule
Benchmark Prompt Testing & \checkmark & \checkmark & \xmark & \checkmark & \xmark \\
Binary Classifier on Hidden States & \checkmark & \checkmark & \checkmark & \checkmark & \xmark \\
Cross-provider Consistency Check & \xmark & \xmark & \xmark & \checkmark & \xmark \\
\hline
\textbf{\texttt{SVIP} (Ours)} & \checkmark & \checkmark & \checkmark & \checkmark & \checkmark \\
\bottomrule
\end{tabular}
}
\end{table*}

\section{Problem Statement}
\label{sec: problem_statement}

The verifiable inference problem involves three parties, as illustrated in Figure \ref{fig:problem_setting}:
\vspace{-6pt}
\begin{enumerate}[wide, labelwidth=!, labelindent=0pt]
    \item \textbf{User:} An individual who lacks sufficient computing resources and seeks to perform expensive LLM inference tasks on given prompts using decentralized computing providers at a low cost.
    \item \textbf{Computing Provider:} Decentralized entities, often \textit{small companies or individuals}, that rent out computational power at competitive prices.
    \item \textbf{Platform:} A commercial platform that profits by connecting users and computing providers.  Importantly, the platform \textit{itself does not require significant computational resources}, as its primary role is to facilitate and monitor the utilization of computational resources from decentralized providers. 
\end{enumerate}

\vspace{-0.4cm}

\paragraph{Threat Model} To reduce costs, a computing provider may not actually use the LLM the user specifies. Instead, it may substitute a significantly smaller model, which returns an inferior result. It may also attempt to evade detection by actively concealing dishonest behavior.

\vspace{-0.4cm}

\paragraph{The Incentive and Goal of Verifiable Inference} To address this threat, platforms are \textbf{commercially incentivized} to maintain user trust by monitoring provider behavior, ensuring that providers cannot cheat. Trust in the platform underpins its \textbf{reputation and business model}; if users cannot trust that model inference is faithfully executed, they are likely to abandon the platform, leading to \textbf{significant financial and operational losses}.

To mitigate this risk, the platform designs and implements a verification protocol that allows users to verify, with high confidence, whether the computing provider used the specified model for inference. Note that during deployment, the protocol should operate \textit{primarily} between the user and the provider, with minimal platform involvement. A satisfactory protocol should meet the following criteria: \textbf{(1) Low False Negative Rate (FNR)}: The protocol should minimize cases where the computing provider \textit{did} use the specified LLM for inference but is incorrectly flagged as not using it. \textbf{(2) Low False Positive Rate (FPR)}: The protocol should rarely confirm that the computing provider used the specified LLM if it actually used another model. \textbf{(3) Efficiency:} The verification protocol should be computationally efficient and introduce minimal overhead for both the computing provider and the user. \textbf{(4) Preservation of Completion Quality:} The protocol should not compromise the quality of the prompt completion returned by the computing provider.
\textbf{(5) Robustness:} The protocol should maintain low FNR and FPR even against adversarial providers attempting to evade detection. 

\paragraph{Necessity of a Lightweight Third-Party Platform.}

We point out that a lightweight third-party platform is \textbf{necessary} for making verification feasible. 
If each user had to prepare and run the full verification protocol locally, the computational and operational cost would be prohibitive. 
A shared platform can perform these tasks once and reuse the verification infrastructure across all users, greatly reducing per-user overhead. 
This design is also consistent with many deployed cryptographic systems, where a small third party plays a trust-anchoring role, such as certificate authorities in public-key infrastructures (TLS/HTTPS), key management or attestation services for trusted execution environments, and auditors in secure multiparty computation or blockchain-based financial systems~\citep{cooper2008internet,rescorla2001ssl,costan2016intel,bonneau2015sok,aggarwal2021cryptocurrencies}.

\subsection{Simple Approaches Do Not Meet All the Criteria}
Table \ref{tab:naive_approaches} evaluates several straightforward approaches for verifiable LLM inference, as well as our proposed protocol (\textsection\ref{sec:method_all}), against the five criteria. All naive approaches fail to satisfy at least one criterion, underscoring the necessity of our method. We exclude solutions that involve multiple computing providers (\textit{e.g.}, cross-verifying results across providers) because such approaches significantly increase user costs, making them impractical for widespread adoption.
\vspace{-0.4cm}

\paragraph{Benchmark Prompt Testing} The user curates a small set of prompt examples from established benchmarks and sends them to the computing provider. If the provider's performance significantly deviates from the reported benchmark metrics for the specified model, the user may suspect dishonest behavior. However, a malicious provider can easily bypass this method by detecting known benchmark prompts and selectively applying the correct model only for those cases, while using an alternative model for all other prompt queries. Additionally, testing such benchmark prompts also increases the user's inference costs.
\vspace{-0.4cm}

\paragraph{Binary Classifier on Hidden States} The user can request the computing provider to return hidden representations from the LLM used for inference, and train a binary classifier on these representations to verify if they come from the specified model. 
However, a simple attack involves the provider caching hidden representations from the correct model that are unrelated to the user's input. The dishonest provider could then use a smaller LLM for inference and return these cached irrelevant representations to deceive the classifier while saving costs. 
\vspace{-0.4cm}

\paragraph{Cross-provider Consistency Check.} Another naive idea is to send the same query to two different providers and flag a provider as suspicious if the responses are not identical. However, this strategy is fundamentally unsatisfactory. First, it at least doubles the user’s inference cost and therefore scales poorly in realistic workloads. Second, even two honest providers running the same specified model need not return identical completions due to sampling stochasticity, implementation details, or hardware differences. Moreover, collusion cannot be ruled out, even in decentralized markets. Multiple providers may be controlled by the same entity or coordinate with each other, allowing them to return matching outputs without ever using the specified model.


\section{Methodology}
\label{sec:method_all}
\begin{figure*}[t!]
    \setlength{\abovecaptionskip}{0.5cm}
    \setlength{\belowcaptionskip}{-0.5cm}
    \centering
    \includegraphics[width=0.75\linewidth]{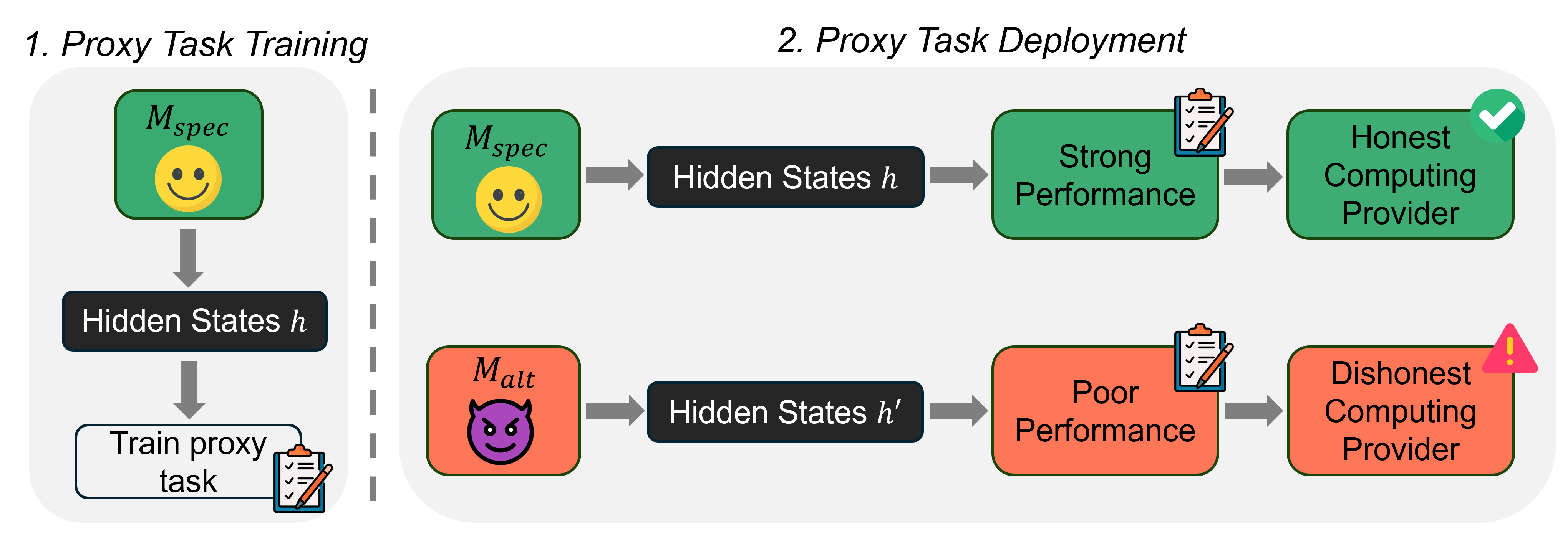}
    \caption{Illustration of the motivation behind our framework. The proxy task is trained solely on hidden states from the specified LLM $M_{spec}$. During deployment, strong performance on the proxy task indicates that the provider used $M_{spec}$ as specified, while poor performance suggests otherwise.}
    \label{fig:motivation}
\end{figure*}


\paragraph{Motivation} It is often challenging to verify whether a computing provider is using an alternative LLM for inference based solely on the returned completion text\footnote{To empirically demonstrate this, we train a binary classifier to distinguish between output texts from a specified model (LlaMA-2–13B) and six smaller alternatives. Using 90,000 prompts for training and 10,000 for testing, the classifier (BERT-base-uncased) achieves an FNR of 36.1\% and an FPR of 58.9\%.}. Our framework addresses this by requiring the computing provider to return not only the generated text but also the processed \textbf{hidden state representations} from the LLM inference process.

\vspace{-0.1cm}

We design and train a \textbf{proxy task} specifically to perform well \textit{only} on the hidden representations generated by the specified model during the protocol's training stage. The intuition behind is that the proxy task transforms the hidden representations into a unique identifier for the model. During deployment, the user can evaluate the performance of the returned hidden states on the proxy task. Strong performance on the proxy task indicates that the correct model was used for inference, while poor performance suggests otherwise. Figure \ref{fig:motivation} provides an illustration.

\vspace{-0.1cm}

Our approach does not depend on expensive cryptographic proofs or protocols, and is highly efficient. Furthermore, it does not involve retraining or fine-tuning the LLMs, operates independently of the model publisher, and can be applied to any LLM with publicly available weight parameters, making it widely
applicable. 

\vspace{-0.3cm}

\paragraph{Notations} Let $x \in \mathcal{V}^*$ denote the prompt query from user, where $\mathcal{V}^*$ represents the set of all possible string sequences for a vocabulary set $\mathcal{V}$. The specified LLM and alternative LLM are denoted as $\mathcal{M}_{spec}$ and $\mathcal{M}_{alt}$, respectively. 

\subsection{A Simple Protocol Based on Hidden States}
\label{sec: simple_protocol}

\paragraph{Protocol Overview} For any LLM $\mathcal{M}$, let $h_\mathcal{M} (x) \in \mathbb{R}^{L\times d_\mathcal{M}}$ represent the \textbf{last-layer} hidden representations of $x$ produced by $\mathcal{M}$, where $L$ is the length of the tokenized input $x$, and $d_\mathcal{M}$ denotes the hidden dimension of $\mathcal{M}$. The computing provider receives $x$ from the user, runs $\mathcal{M}$, and returns $h_{\mathcal{M}}(x)$ to user for subsequent verification. However, to reduce the size of the hidden states returned, we additionally apply a proxy task feature extractor network $g_{\theta}(\cdot): \mathbb{R}^{L \times d_\mathcal{M} }\rightarrow \mathbb{R}^{d_g}$ parameterized by $\theta$, where $d_g$ represents the proxy task feature dimension. The computing provider now also runs $g_{\theta}(\cdot)$ and returns a compressed vector
  $  z(x) := g_{\theta}(h_\mathcal{M}(x)) $
 of dimension $d_g$ to the user, significantly reducing the communication overhead. Specifically, for each prompt query, the compressed vector only takes approximately $4$ KB when $d_g$ is set to $1024$.

\vspace{-0.1cm}

The user is required to perform two tasks locally: obtaining the predicted proxy task output and the label. First, the user runs $f_{\phi}(\cdot)$, using the returned proxy task feature $z(x)$ as input to compute $f_{\phi}(z(x))$. Here, $f_{\phi}(\cdot): \mathbb{R}^{d_g} \rightarrow \mathcal{Y}$ is the proxy task head parameterized by $\phi$, where $\mathcal{Y}$ denotes the label space. Second, the user applies a labeling function for the proxy task. We adopt a self-labeling function $y(x): \mathcal{V}^{*} \rightarrow \mathcal{Y}$, which derives the label directly from the input, eliminating the need for external labels or specialized annotators\footnote{For instance, we can define $y(x)$ as the Set-of-Words (SoW) representation of the input $x$, which captures the presence of each word in a fixed vocabulary, regardless of frequency. As a concrete example, if $\mathcal{V} = \{a, b, c, d\}$ and $x = ``abcc"$, the SoW label $y(x)$ would be a four-dimensional vector $(1, 1, 1, 0)$, indicating whether each token in $\mathcal{V}$ appears in $x$.}. The label can either be a scalar or a vector.

\vspace{-0.1cm}

Finally, the user checks whether $f_{\phi}(z(x))$ matches $y(x)$. Our training process below ensures that, with high probability, $f_{\phi}(z(x)) = y(x)$ when $\mathcal{M}_{spec}$ is used for inference, and that this does not hold for other models, as the proxy task is exclusively trained on the hidden representation distribution induced by $\mathcal{M}_{spec}$. This completes our protocol. Refer to Figure \ref{fig:protocol_simple} for a detailed illustration.

\vspace{-0.3cm}

\paragraph{Proxy Task Training}  With a properly defined loss function $\ell: \mathcal{Y} \times \mathcal{Y} \rightarrow \mathbb{R}$ and a training dataset $\mathcal{D}$, the platform trains the proxy task according to the following objective: 
\begin{align*}
    \phi^{*}, \theta^{*} = \arg\min_{\phi, \theta} \mathbb{E}_{x \sim \mathcal{D}} \left[ \ell \left( f_{\phi}( g_{\theta}(h_{\mathcal{M}_{spec}}(x))), y(x) \right) \right].
    \label{eqn:training_no_secret} 
\end{align*}

\vspace{-0.5cm}

\paragraph{Protocol Deployment}  With the optimized $\phi^*$ and $\theta^*$, we define the \textbf{verification function} as 
 $   V(x, z(x); \phi^{*}, \theta^{*}) = \mathds{1}\left(f_{\phi^{*}}(z(x) ) = y(x)\right)$, where $z(x) = g_{\theta^{*}}(h_{\mathcal{M}}(x))$ is returned by the computing provider. If the value of the verification function is $1$ (or $0$), we conclude that the computing provider is indeed (or is not) using $\mathcal{M}_{spec}$ for inference with high probability. Now, the low FNR and low FPR criteria introduced in Section \ref{sec: problem_statement} can be formally expressed as follows:
\begin{equation}
\begin{split}
    \quad &\textbf{Low FNR}: \mathbb{P}\left( V(x, z(x); \phi^{*}, \theta^{*} ) = 0 ~\middle|~ 
    \mathcal{M}_{spec}~\text{used}\right) \leq \alpha; \\
    \quad &\textbf{Low FPR}: \mathbb{P}\left( V(x, z(x); \phi^{*}, \theta^{*})  = 1 ~\middle|~ 
    \mathcal{M}_{spec}~\text{\textit{not} used}\right) \leq \beta.\notag
\end{split}
\label{eqn: FNR and FPR}
\end{equation}

 \vspace{-0.2cm}

Here $\alpha$ and $\beta$ are predefined thresholds. 

\paragraph{A Hypothesis Testing Framework for a Single Final Conclusion} In practical scenarios, conclusions about a computing provider’s honesty are based on \textbf{multiple distinct} prompt queries rather than a single one. Each prompt query yields a binary acceptance decision (pass or fail), and the final judgment is made by examining the empirical fraction of passes across all queries. This naturally leads to a classical hypothesis testing view, where the null hypothesis corresponds to an honest provider and the alternative to a dishonest one, enabling multiple query outcomes to be aggregated into a \textit{single final} decision. 

The following theorem shows that, once single-query error rates (FNR/FPR) are controlled, aggregating over a modest number of queries already makes the final decision highly reliable. For example, with per-query FNR and FPR below $5\%$, aggregating even $B = 30$ queries drives both types of errors below $10^{-5}$, making the final decision effectively dependable in practice. Refer to Appendix \ref{sec:app_hypothesis} for detailed analysis and derivation.

\begin{theorem}
\label{thm:hypothesis-testing}
Suppose the protocol has per-query false negative rate $\mathrm{FNR}$ and false positive rate $\mathrm{FPR}$. 
Let $V_1,\dots,V_B \in \{0,1\}$ be the verification outcomes of $B$ independent queries, and let $\bar V = \frac{1}{B}\sum_{i=1}^B V_i .$
Consider any decision rule that declares the provider honest whenever $\bar V \ge \tau$, for a threshold $\tau$ satisfying 
$
\mathrm{FPR} < \tau < 1 - \mathrm{FNR}.
$
Then both the type-I error (incorrectly rejecting an honest provider) and the type-II error (failing to detect a dishonest provider) decrease exponentially fast in $B$.
\end{theorem}



    

\subsection{\texttt{SVIP}: A Secret-based Protocol for Verifiable LLM Inference }
\label{sec:methods_secret}

\paragraph{From Simple Protocol to Secret-based Protocol} The simple protocol, despite its strong potential in discriminating whether the specified model is actually used, is vulnerable to malicious attacks from the computing provider. A dishonest provider may attempt to bypass the verification process without running $\mathcal{M}_{spec}$. Since all the provider needs to return is a vector of dimension $d_g$, an attacker could adversarially optimize a vector $\Tilde{z} \in \mathbb{R}^{d_g}$ directly, without actually running $g_{\theta^*}(\cdot)$ and using \textit{any} LLM. We refer to this as a \textbf{direct vector optimization attack}. Specifically, if the self-labeling function is public, the adversary can run the labeling function $y(x)$ themselves for each input $x$ and then directly find $\Tilde{z}$ so that  
\begin{align}
  \Tilde{z}^* = \arg\min_{\Tilde{z}}  \ell \left( f_{\phi^{*}}(\Tilde{z}), y(x) \right).  
  \label{eqn:adversarial}
\end{align}

\vspace{-0.2cm}

Ultimately, $\Tilde{z}^*$ is returned to the user to deceive the verification protocol. As shown in Appendix \ref{sec:ablation_study}, this attack achieved an attack success rate (ASR) of $99.90\%$, indicating that the protocol's security requires further enforcement.

To strengthen the protocol's security, we introduce a ``secret" mechanism. A complete illustration is provided in Figure \ref{fig:protocol_secret}. Particularly, the platform assigns a ``secret'' $s \in \mathcal{S}$ exclusively  to the user, which is never shared with the computing provider. Here, $\mathcal{S}$ represents the secret space. For example, $\mathcal{S}$ can be defined as the space of $d_s$-dimensional binary vectors, represented as $\{0, 1\}^{d_s}$.

\vspace{-0.2cm}

\paragraph{Introducing Secret into the Self-labeling Function}
The self-labeling function with secret is now defined as $y(x, s): \mathcal{V}^{*} \times \mathcal{S} \rightarrow \mathcal{Y}$. The property below is essential for an ideal self-labeling function.

\begin{property}[Secret Distinguishability]
 For the same input $x$, given two different secrets $s' \neq s$, with a pre-defined lower-bound probability $\delta$, the resulting labels should be different with high probability:  \begin{align}
    \mathbb{P}(y(x, s)\neq y(x,s')) \geq \delta.\notag
\end{align}
If $\mathcal{Y} $ is a continuous space, with a pre-defined threshold, this property is equivalent to:\begin{align}
       \mathbb{P}(\| y(x, s) -y(x,s')\|_2   \geq \text{threshold}) \geq \delta.
 \label{eqn_label_property_continuous}
 \end{align}
\label{Property_1}
\end{property}\vspace{-22pt}

Property \ref{Property_1} ensures that a malicious computing provider, without access to the specific $s$, cannot determine or naively guess the true label, thus rendering the direct vector optimization attack ineffective. Meanwhile, the user, with knowledge of $s$, can still compute the correct label.

A simple rule-based self-labeling function (\textit{e.g.}, the SoW representation) cannot ensure that Property \ref{Property_1} holds. To enforce this property, we introduce a trainable labeling network $y_{\gamma}(x,s): \mathcal{V}^* \times \mathcal{S} \rightarrow \mathcal{R}^{d_y}$ parameterized by $\gamma$, which takes $x \in \mathcal{V}^*$ and $s \in \mathcal{S}$ as input and outputs a continuous label vector of dimension $d_y$. This network is trained with the following contrastive loss:
\begin{align}
\gamma^* = \arg\min_{\gamma}   -\mathbb{E}_{x \sim \mathcal{D}, s, s' \sim \mathcal{S}} \left[ \| y_{\gamma}(x, s) - y_{\gamma}(x, s') \|_2 \right].\notag
\label{label_network_loss_func}
\end{align}
\vspace{-0.3cm}
\paragraph{Introducing Secret into the Proxy Task}
Once the labeling network is optimized, we also need to include the secret $s$ into the proxy task. Our design is to embed $s$ as a task token using a secret embedding network (\textit{e.g.}, an MLP), denoted as $t_{\psi}(s): \mathcal{S}\rightarrow \mathbb{R}^{d_\mathcal{M}}$, parameterized by $\psi$. Note that this secret embedding network $t_{\psi}(s)$ is only kept to the platform. Then, the platform distributes $t_{\psi}(s)$ to the computing provider, who concatenates $t_\psi(s)$ with $h_{\mathcal{M}}(x)$, 
 runs $g_{{\theta}}(\cdot)$, and returns $z(x) = g_{{\theta}}(t_\psi(s) \oplus h_{\mathcal{M}}(x))$, where $\oplus$ denotes concatenation. 

The training objective is now modified by incorporating randomly sampled secrets during training: 
{\small
\begin{align}
    &\phi^{*}, \theta^{*}, \psi^{*} = \notag \\
    &\arg\min_{\phi, \theta, \psi} \mathbb{E}_{x \sim \mathcal{D}, s \sim \mathcal{S}} \left[ \ell \left( f_{\phi}(g_{\theta}(\underline{\underline{t_{\psi}(s)}}
\oplus h_{\mathcal{M}_{spec}}(x))), y_{\gamma^*}(x,\underline{\underline{s}}
) \right) \right]. \notag
\end{align}
}
As before, the user receives $z(x)$ from the computing provider. However, now that $\mathcal{Y}$ is a continuous space, a threshold $\eta$ is required to determine whether the predicted proxy task output $f_{\phi^{*}}(z(x) )$ matches the label vector $y_{\gamma^*}(x,s)$. Specifically, $f_{\phi^{*}}(z(x) )$ is considered a match to $y_{\gamma^*}(x,s)$ if the $L_2$ distance between them is below the pre-defined threshold $\eta$,  indicating $\mathcal{M}_{spec}$ was actually used:
\begin{align}
    V(x, z(x); \phi^{*}, \theta^{*}, \psi^{*}) = \mathds{1}\left(\|f_{\phi^{*}}(z(x) ) - y_{\gamma^*}(x,s)\|_2 \leq \eta \right).\notag
\end{align}

In practice, we propose setting the threshold based on the \textbf{conditional empirical distribution} of $d(x,s) := \|f_{\phi^{*}}(z(x) ) - y_{\gamma^*}(x,s)\|_2$, given that $\mathcal{M}_{spec}$ is used for inference. We select the upper 95th percentile to ensure a FNR of $5 \%$. 

\subsection{Security Analysis}
\label{sec:attacks}
As previously discussed, the direct vector optimization attack described in Eq.~(\ref{eqn:adversarial}) is no longer feasible due to the introduction of the secret mechanism. In this section, we discuss other potential attacks as a security analysis towards our protocol. Additional possible attacks are discussed in Appendix \ref{app_sec:attacks}.

\vspace{-0.2cm}

\paragraph{Adapter Attack Under Single Secret} A malicious attacker could attempt an adapter attack if they collect enough prompt samples  $\mathcal{D}' = \{x_i \}^M_{i=1}$ under a \textit{single} secret $s$. The returned vector from an honest computing provider should be $z(x) = g_{{\theta}^*}(t_{\psi^*}(s) \oplus h_{\mathcal{M}_{spec}}(x))$. The attacker's goal is to train an adapter that mimics the returned vector, but by using an alternative LLM, $\mathcal{M}_{alt}$.

To this end, we define the adapter  $a_{\lambda}(\cdot): \mathbb{R}^{d_{\mathcal{M}_{alt}}} \rightarrow \mathbb{R}^{d_{\mathcal{M}_{spec}}}$, parameterized by $\lambda$, which transforms the hidden states of $\mathcal{M}_{alt}$ to approximate those of $\mathcal{M}_{spec}$. The returned vector is then
$g_{{\theta}^*}(t_{\psi^*}(s) \oplus a_{\lambda}(h_{\mathcal{M}_{alt}}(x)))$. The attacker's objective is to minimize the $L_2$ distance between the returned vector generated by $\mathcal{M}_{spec}$ and the vector produced by $\mathcal{M}_{alt}$ with the adapter:
\vspace{0.1cm}
\begin{align}
  \lambda^*=  \arg\min_{\lambda} \mathbb{E}_{x \sim \mathcal{D}'} \notag \|  g_{{\theta}^*}(t_{\psi^*}(s) \oplus h_{\mathcal{M}_{spec}}(x))
\\-g_{{\theta}^*}(t_{\psi^*}(s) \oplus a_{\lambda}(h_{\mathcal{M}_{alt}}(x)))
  \|_2.
\end{align}
By minimizing this objective, the attacker seeks to make the output of $\mathcal{M}_{alt}$ with the adapter indistinguishable from that of $\mathcal{M}_{spec}$, effectively bypassing the protocol. Once the adapter is well-trained, as long as the secret $s$ remains unchanged, the attacker can rely solely on $\mathcal{M}_{alt}$ in future verification queries without being detected.

\vspace{-0.2cm}

\paragraph{Secret Recovery Attack Under Multiple Secrets} The secret mechanism is enforced by distributing the secret $s$ to the user, while only providing the secret embedding $t_{\psi^*}(s) $ to the computing provider. However, a sophisticated computing provider may attempt to recover the original secret by posing as a user and collecting multiple secrets and corresponding embeddings. 
A straightforward approach would involve recovering $s$ from $t_{\psi^*}(s) $, thereby undermining the secret mechanism.

Suppose the attacker has curated a dataset of secret-embedding pairs, $D_{\text{secret}} = \{s_j, t_{\psi^*}(s_j)\}^N_{j=1}$. The attacker could then train an inverse model $i_{\rho}:\mathbb{R}^{d_\mathcal{M}} \rightarrow \mathcal{S}$, parameterized by $\rho$, to map the secret embedding back to the secret space. If $\mathcal{S}$ is continuous, the training objective can be formalized as:
\vspace{-1pt}
\begin{align}
    \rho^{*} = \arg\min_{\rho} \mathbb{E}_{ s \sim \mathcal{D}_{\text{secret}}} \| 
    i_{\rho}(t_{\psi^*}(s)) - s
    \|_2.\notag
    \label{eqn:secret_attack}
\end{align}

\vspace{-8pt}
Once the inverse model is optimized, the true label $y(x,s)$ again becomes accessible to the malicious provider. Consequently, the secret-based protocol effectively collapses to the simple protocol without secret protection, leaving it vulnerable to the direct vector optimization attack.

\vspace{-0.2cm}

\paragraph{Defense: The Update Mechanism} To defend against the attacks discussed above, we propose a simple update mechanism: \textbf{(1)} In defense of the adapter attack, once the prompt queries for a given secret reach a pre-defined threshold $M^*$, the next secret is activated. Meanwhile, we enforce a limit on how often the next secret can be activated, preventing attackers from acquiring too many secrets within a short period. \textbf{(2)} When a total of $N^*$ secrets have been used, the entire protocol should be retrained by the platform\footnote{Specifically, this retraining can be performed using a different random seed and training recipe. As shown in Section \ref{sec:cost_analysis}, the retraining process is efficient.}. In practice, the values of $M^*$ and $N^*$ can be determined empirically, as discussed in Section \ref{sec:results_attack}. 

\section{Experiments}
\label{sec:experiments}



\paragraph{Experiment Setup}
To simulate realistic LLM usage scenarios, we primarily use the \texttt{LMSYS-Chat-1M} conversational dataset \citep{zheng2023lmsyschat1m}, which consists of one million real-world conversations. Results on additional datasets are provided in Appendix~\ref{sec:app_unseen}. For the models,
we select $5$ widely-used LLMs as the specified models, ranging in size from $13$B to $70$B parameters and spanning multiple model families. As alternative models, we use $6$ smaller LLMs, each with parameters up to $7$B. Refer to Appendix \ref{sec:app_data_model} for details.
The labeling network $y_\gamma(\cdot)$
uses a pretrained sentence transformer \citep{reimers2019sentence} to embed the text input $x$ and an MLP to embed the secret $s$, where $s \in \{0,1\}^{d_s}$ and $d_s$ is set to $48$. The outputs of both embeddings are concatenated and passed through another MLP to produce a continuous label vector of $128$ dimensions.
The proxy task feature extractor $g_\theta(\cdot)$ is a $4$-layer transformer, while both the proxy task head $f_\phi(\cdot)$ and task embedding network $t_\psi(\cdot)$ are implemented as MLPs. Full details can be found in Appendix \ref{sec: app_protocol_training}.

\begin{table*}[t!]
\caption{FNR and FPR across different specified models on the test dataset of \texttt{LMSYS-Chat-1M}. Our protocol keeps FNR under $5\%$ and FPR under $3 \%$ across all scenarios. We implement a \texttt{Random} baseline where the computing provider generates random hidden representations directly without using any LLM. }
\label{tab_accuracy_test}
\centering
\resizebox{0.85\linewidth}{!}{
\small 
\begin{tabular}{c|c|ccccccc}
\toprule[1.5pt]
\multirow{2}{*}{\textbf{Specified Model}} & \multirow{2}{*}{\textbf{FNR $\downarrow$}}& \multicolumn{7}{c}{\textbf{FPR $\downarrow$}} \\ 
 &  & \texttt{Random} & \texttt{GPT2-XL} & \texttt{GPT-NEO-2.7B} & \texttt{GPT-J-6B} & \texttt{OPT-6.7B} & \texttt{Vicuna-7B} & \texttt{Llama-2-7B} \\
\midrule
\texttt{Llama-2-13B}     & 4.41\%  & 1.97\%  & 1.90\%  & 1.77\% & 1.75\% & 2.03\% & 2.44\% & 2.04\% \\
\texttt{GPT-NeoX-20B}    & 3.47\%  & 0.00\%  & 0.00\%  & 0.00\% & 0.00\% & 0.00\% & 0.00\% & 0.00\% \\
\texttt{OPT-30B}         & 3.42\%  & 0.05\%  & 0.33\%  & 0.61\%  & 0.47\% & 0.83\% & 0.34\% & 0.35\% \\
\texttt{Falcon-40B}      & 3.02\%  & 0.00\%  & 0.00\%  & 0.01\% & 0.00\% & 0.00\% & 0.00\% & 0.00\% \\
\texttt{Llama-3.1-70B}   & 3.13\%  & 0.26\%  & 1.97\%  & 1.04\% & 1.98\% & 2.07\% & 0.90\% & 0.81\% \\
\bottomrule[1.5pt]
\end{tabular}}
\end{table*}

\subsection{Results of Protocol Accuracy}
\label{sec:results_accuracy}
We evaluate the accuracy of our protocol by examining the empirical estimate of FNR and FPR, as outlined in Eq.~(\ref{eqn:estimate_fnr_fpr}). To apply the verification function, we first determine the threshold $\eta$ on a validation dataset during proxy task training. We then evaluate the empirical FNR and FPR on a \textit{held-out} test dataset with $10,000$ samples. For each test prompt, we pair it with $30$ randomly sampled secrets to ensure a reliable evaluation result. For FPR calculations, we simulate scenarios where the computing provider uses an alternative, smaller LLM to produce the hidden representations, and applies $g_{\theta^*}(\cdot)$ on those outputs.

As shown in Table \ref{tab_accuracy_test}, \texttt{SVIP} consistently achieves low FNR and FPR \textbf{for individual queries} across all specified LLMs. The FNR remains below $5\%$ per query, indicating that our protocol rarely falsely accuses an honest computing provider. Moreover, when faced with a dishonest provider, the FPR stays under $3\%$ per query regardless of the alternative model employed, highlighting the protocol's strong performance in detecting fraudulent behavior. We further demonstrate the generalizability of our protocol to unseen datasets in Appendix \ref{sec:app_unseen}.

Figure \ref{fig: d_distribution} shows the empirical \textit{test} distribution of $d(x,s)$, the $L_2$ distance between the predicted proxy task output and the label vector, under different model usage scenarios. The clear separation in the distributions provides strong evidence for the high accuracy of \texttt{SVIP}: when the specified model is actually used, $d(x,s)$ is significantly smaller compared to when an alternative model is used.

\begin{figure}[t!]
    \centering
    \includegraphics[width=0.70\columnwidth]{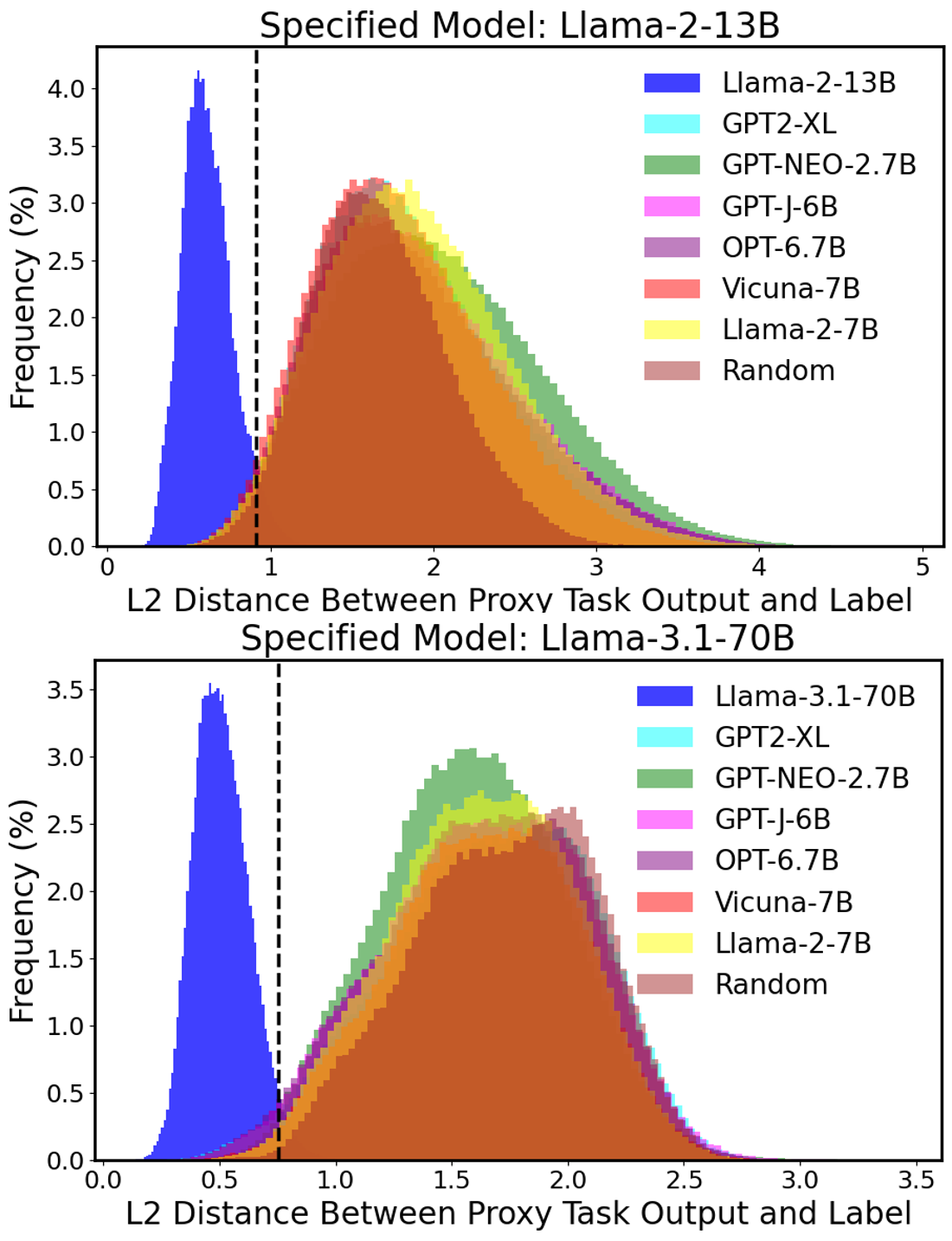}
    \caption{Empirical distribution of the $L_2$
    distance between the predicted proxy task output $f_{\phi^{*}}(z(x) )$  and the label vector $y_{\gamma^*}(x,s)$ on the test dataset of \texttt{LMSYS-Chat-1M}. Each figure corresponds to a different specified model. The distributions compare the $L_2$ distances when the specified model is used versus various alternative models. The clear separation between the distributions, marked by the vertical threshold line, ensures the high accuracy of our protocol in distinguishing between correct and incorrect model usage. More examples can be found in \ref{sec:app_unseen}.
    }
    \label{fig: d_distribution}
\end{figure}

\begin{table*}[t]
\caption{Attack Success Rate for the secret recovery attack, presented as a function of the number of secret-embedding pairs collected. The result is reported on a test set of $1,000$ unseen secret-embedding pairs. The ASR remains below $50\%$ even after collecting $200,000$ pairs.}
\label{tab: secret_recovery}
\centering
\resizebox{0.7\linewidth}{!}{
\small 
\begin{tabular}{c|cccccccc}
\toprule[1.5pt]
\textbf{Specified Model} & \textbf{1,000} & \textbf{5,000} & \textbf{10,000} & \textbf{50,000} & \textbf{100,000} & \textbf{200,000} & \textbf{500,000} & \textbf{1,000,000} \\
\midrule
\texttt{Llama-2-13B}     & 0.0\%   & 0.0\%  & 0.0\%  & 2.7\%  & 5.8\%   & 30.1\%  & 65.1\%  & 69.5\% \\
\texttt{GPT-NeoX-20B}    & 0.0\%   & 0.0\%  & 0.0\%  & 0.0\%  & 1.2\%   & 19.6\%  & 30.4\%  & 59.9\% \\
\texttt{OPT-30B}         & 0.0\%   & 0.0\%  & 0.0\%  & 1.2\%  & 6.4\%   & 40.1\%  & 84.6\%  & 92.3\% \\
\texttt{Falcon-40B}      & 0.0\%   & 0.0\%  & 0.0\%  & 0.1\%  & 2.9\%   & 12.4\%  & 40.7\%  & 72.9\% \\
\texttt{Llama-3.1-70B}   & 0.0\%   & 0.0\%  & 0.0\%  & 0.5\%  & 3.6\%   & 17.3\%  & 21.3\%  & 84.9\% \\
\bottomrule[1.5pt]
\end{tabular}}
\end{table*}

As discussed earlier, aggregating verification results over multiple distinct prompt queries enables the user to reach a single final decision with high confidence. As an illustrative case, when using \texttt{Llama-3.1-70B} as the specified model and \texttt{Llama-2-7B} as the alternative, we achieve an FPR of $0.81\%$ and an FNR of $3.13\%$. With a properly chosen decision threshold, the type-I error rate (incorrectly flagging an honest provider as dishonest) and type-II error rate (failing to detect a dishonest provider) rates are $1.7 \times 10^{-49}$ and $0.0$, respectively. 



\subsection{Computational Cost Analysis of the Protocol}
\label{sec:cost_analysis}

Table \ref{tab: deployment_cost} details the runtime per prompt query and GPU memory consumption during the deployment stage. Across all specified models, the verification process takes under $0.01$ seconds per prompt query for both the computing provider and the user. For example, verifying the \texttt{Llama-2-13B} model for each prompt query takes only $0.0017$ seconds for the computing provider and $0.0056$ seconds for the user, in stark contrast to zkLLM \citep{sun2024zkllm}, where generating a single proof requires $803$ seconds and verifying the proof takes $3.95$ seconds for the same LLM. The proxy task feature extractor $g_\theta(\cdot)$, run by the computing provider, consumes approximately $980$ MB of GPU memory, imposing only minimal overhead. On the user side, the proxy task head $f_\phi(\cdot)$ and labeling network $y_\gamma(\cdot)$ require a total of $1428$ MB, making it feasible for users to run on local machines without high-end GPUs. Additionally, we record the required proxy task retraining time in Table \ref{tab: training_time}. 
Overall, retraining the proxy task takes less than $1.5$ hours on a single GPU,  
allowing for efficient protocol update.


\vspace{-2.0pt}

\subsection{Results of Protocol Security}
\label{sec:results_attack}

\paragraph{Robustness Evaluation Against Adapter Attack} To simulate the adapter attack, we assume an attacker collects a dataset of size $M$, consisting of prompt samples associated with a single secret $s$. The attack is considered successful if the resulting adapter passes the verification function \textbf{when secret $s$ is applied}. Additional details about the experimental setup can be found in Appendix \ref{sec:app_adapter}.

As shown in Figure \ref{fig:adapter}, using a $50 \%$ ASR threshold, \texttt{Llama-2-13B} resist attacks with up to $400$ prompt samples, regardless of the alternative model used. For \texttt{Llama-3.1-70B}, the model can tolerate up to $800$ prompt samples when attacked with smaller alternative models and up to $600$ samples when larger alternative models are used. 

\begin{figure}[t]
    \centering
    \setlength{\belowcaptionskip}{-0.5cm}
    \includegraphics[width=1.0\linewidth]{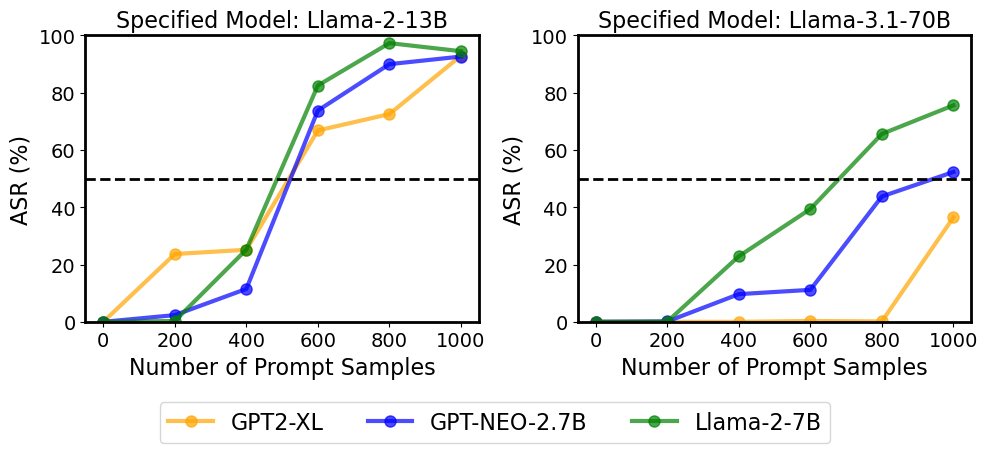}
    \caption{
   Attack Success Rate for the adapter attack, plotted as a function of the number of prompt samples collected \textit{under each single secret}.}
    \label{fig:adapter}
\end{figure}

\vspace{-5pt}
\paragraph{Robustness Evaluation Against Secret Recovery Attack}

We assume the attacker has collected $N$ secret-embedding pairs to train an inverse model to predict the original secret from its embedding. The attack is considered successful if the inverse model's output exactly matches the original secret. Table \ref{tab: secret_recovery} demonstrates the ASR across different specified models as a function of $N$. The attacker is unable to recover any secrets when $N \leq 10,000$. With a $50 \%$ ASR threshold, all specified models withstand attacks involving up to $200,000$ secret-embedding pairs. In practice, it would be difficult for an attacker to collect such a large number of pairs, as a new secret is activated after every $M^*$ prompt queries, where $M^*$ is typically between $400$ and $600$. By setting $N^*$ to $200,000$, \texttt{SVIP} can overall securely handle approximately $80$ to $120$ million prompt queries before a full protocol retraining is needed, demonstrating its robustness against adaptive attack strategies discussed here.

\section{Conclusion}

In this paper, we present \texttt{SVIP}, a novel framework that enables accurate, efficient, and robust verifiable inference for LLMs. We hope that our work will spark further exploration into this area, fostering trust and encouraging wider adoption of open-source LLMs.

\section*{Impact Statement}
\label{app_sec:ethics}
In this work, we address the challenge of verifiable LLM inference, aiming to foster trust between users and computing service providers. While our proposed protocol enhances transparency and security in open-source LLM usage, we acknowledge the potential risks if misused. Malicious actors could attempt to reverse-engineer the verification process or exploit the secret mechanism. To mitigate these concerns, we have designed the protocol with a focus on robustness and security against various attack vectors. Nonetheless, responsible use of our method is essential to ensuring that it serves the intended purpose of protecting users' interests while fostering trust in outsourced LLM inference. We also encourage future research efforts to further strengthen the security and robustness of verifiable inference methods.

\nocite{langley00}

\bibliography{example_paper}
\bibliographystyle{icml2026}

\newpage
\appendix
\onecolumn
\section{Accessibility}
\label{app_sec:code_repo}
Our code repository is available at \url{https://github.com/ASTRAL-Group/SVIP}. In Section \ref{sec:experiments}, we provide a detailed description of the experimental setup, including dataset, models, protocol training details, and evaluation procedures. Additional experimental details can be found in Appendix \ref{sec:app_experiments}.

\section{Discussions}
\subsection{Limitations and Future Work}
\label{app_sec:limitation}


Unlike cryptographic verifiable computation techniques, our approach does not offer a strict security guarantee. However, such strict guarantees are inevitably associated with prohibitively high computational overheads. For example, Zero-Knowledge-based verifiable inference methods (\textit{e.g.}, zkLLM~\citep{sun2024zkllm}) can take over 800 seconds to produce a proof for a single prompt and still require several seconds for verification, far exceeding the latency that real-world LLM services can tolerate. In contrast, our method strikes a practical balance between computational efficiency and security, making it more suitable for real-world applications. Specifically, the protocol’s low single-query FPR and FNR, together with the hypothesis testing framework discussed in Section~\ref{sec: simple_protocol} (Theorem~\ref{thm:hypothesis-testing}), provide a reasonable and practically meaningful level of security when aggregating over multiple queries.

As a potential future direction, we observe that our current design focuses on the prompt-side hidden representation, which already provides sufficient signal for effective verification in our setting. It may nonetheless be of interest to explore whether incorporating representations that jointly encode both the prompt and the specified model’s completion could offer additional security margins under certain threat models. 



\subsection{Protocol Illustration}
\begin{figure}[t!]
    \centering
    \setlength{\belowcaptionskip}{-0.2cm} 
    \begin{subfigure}[t]{0.4\linewidth}
        \centering
        \includegraphics[width=\linewidth]{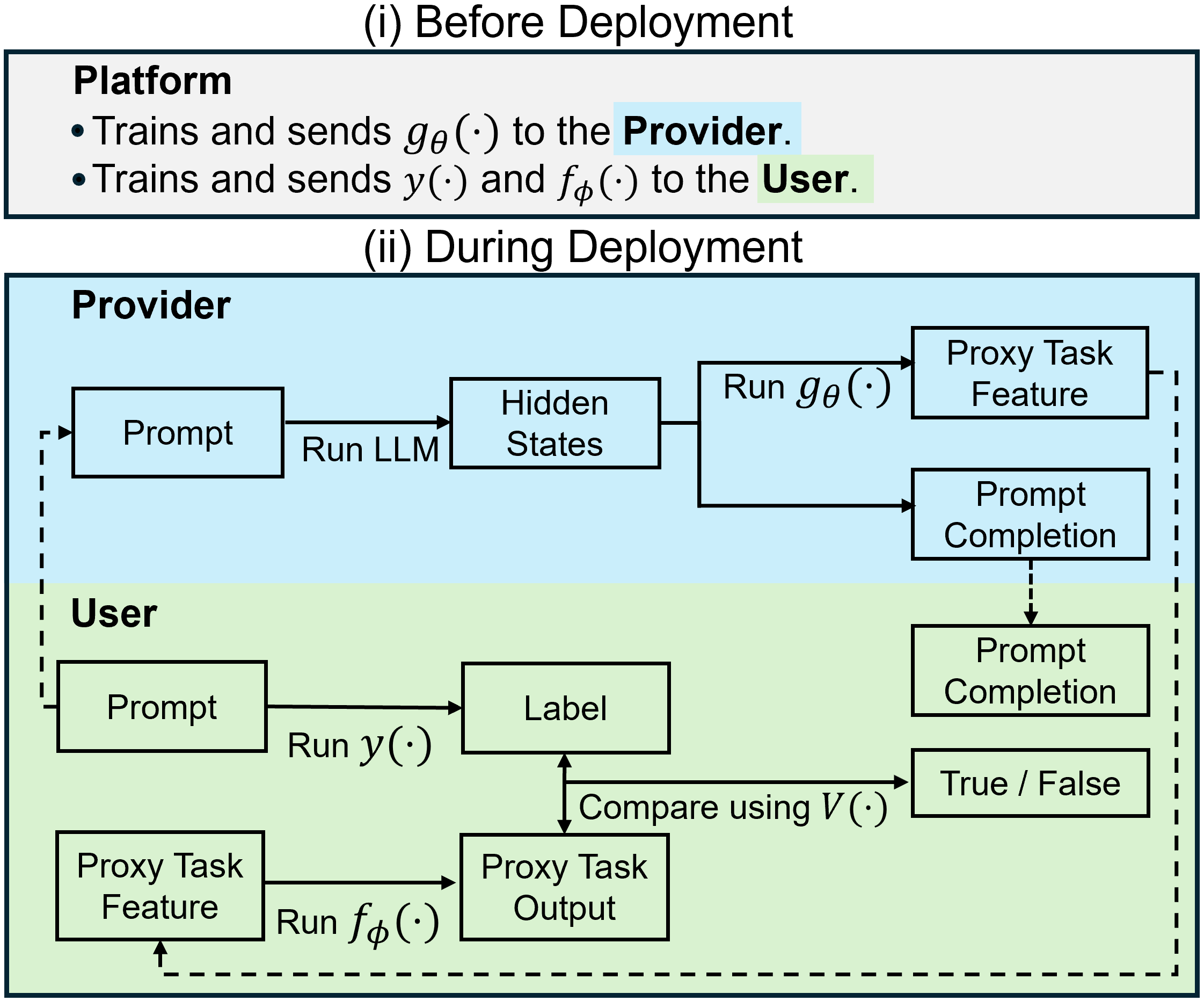}
        \caption{Simple Protocol}
        \label{fig:protocol_simple}
    \end{subfigure}
    \hspace{0.05\linewidth} 
    \begin{subfigure}[t]{0.4\linewidth}
        \centering
        \includegraphics[width=\linewidth]{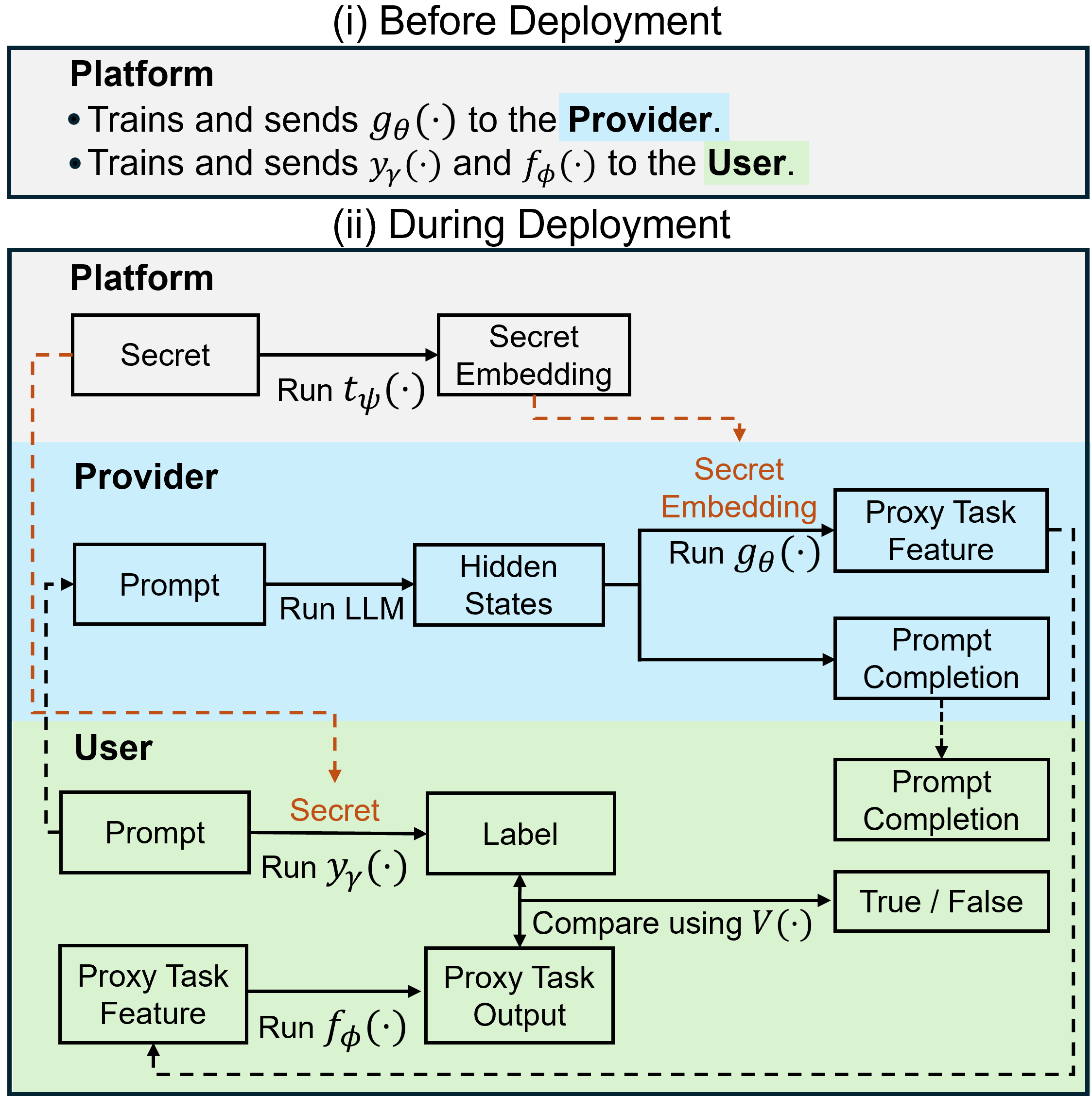}
        \caption{Secret-based Protocol}
        \label{fig:protocol_secret}
    \end{subfigure}
    \caption{Illustration of (a) the simple protocol (Section \ref{sec: simple_protocol}); (b) secret-based protocol (Section \ref{sec:methods_secret}).
    }
    \label{fig:protocol_illustrations}
\end{figure}

To provide an intuitive understanding of our verification protocol, we include a visual illustration in Figure~\ref{fig:protocol_illustrations}. 

\subsection{Hypothesis Testing for Verification Using a Batch of Prompt Queries}
\label{sec:app_hypothesis}
A single prompt query may occasionally yield an incorrect verification result due to FNR or FPR. In practice, users often have multiple prompt queries $\{x_i \}_{i=1}^B$, where $B$ denotes the number of prompts. For each prompt, we observe $V_i := V(x_i, z(x_i); \phi^{*}, \theta^{*}, \psi^{*}) \in \{0,1\}, i \in [B]$.

We formalize this problem as follows: Suppose $Z$ represents whether the computing provider is acting honestly, \textit{i.e.}, the specified model is used, where $Z = 1$ denotes honesty and $Z = 0$ otherwise. When $Z = 1$, $V_i \overset{\text{i.i.d.}}{\sim}
 \text{Bernoulli}(p_1)$. By definition, $p_1$ corresponds to the True Positive Rate (TPR) of our protocol:
\begin{align}
   p_1 = \mathbb{P}(V_i = 1  \mid \mathcal{M}_{\text{spec}} ~\text{is used for inference}) = \text{TPR}.
\end{align}

Similarly, when $Z = 0$, $V_i \overset{\text{i.i.d.}}{\sim}
 \text{Bernoulli}(p_0)$, where $p_0$ is the False Positive Rate (FPR) of our protocol.

In practice, we determine whether the provider is acting honestly based on the mean of the observed values $\{V_i\}_{i=1}^B$, denoted as
\[
\bar{V} = \frac{1}{B} \sum_{i=1}^B V_i.
\]

To achieve a reliable conclusion with high confidence, \textbf{hypothesis testing} can be applied. Specifically, the null hypothesis assumes that the computing provider is acting honestly, \textit{i.e.}, $Z = 1$, and the rejection region is $\bar{V} < \tau$. 

\paragraph{Finite-sample Theoretical Guarantee.}
We now present a finite-sample guarantee for the batch-level verification rule.

\begin{theorem}[Exponential error decay under query aggregation]
\label{thm:appendix_exp_decay}
Let $V_1,\dots,V_B$ be the per-query verification outcomes defined above, and let
\[
\bar V = \frac{1}{B}\sum_{i=1}^B V_i.
\]
Suppose that under honesty ($Z=1$) we have $V_i \overset{\text{i.i.d.}}{\sim}\mathrm{Bernoulli}(p_1)$ and under dishonesty ($Z=0$) we have $V_i \overset{\text{i.i.d.}}{\sim}\mathrm{Bernoulli}(p_0)$, with $p_1 > p_0$.  
Consider the decision rule that declares the provider honest whenever $\bar V \ge \tau$ for some threshold $\tau$ satisfying
\[
p_0 < \tau < p_1.
\]
Then the type-I and type-II error probabilities satisfy
\[
\Pr(\bar V < \tau \mid Z=1) \le \exp\!\bigl(-2B (p_1 - \tau)^2 \bigr),
\qquad
\Pr(\bar V \ge \tau \mid Z=0) \le \exp\!\bigl(-2B (\tau - p_0)^2 \bigr).
\]
In particular, both error probabilities decay exponentially in $B$.
\end{theorem}

\begin{proof}
Under $Z=1$, we have $V_i \overset{\text{i.i.d.}}{\sim} \mathrm{Bernoulli}(p_1)$, so
\[
\mathbb{E}[\bar V \mid Z=1] = p_1.
\]
Because $\tau < p_1$, we can apply Hoeffding’s inequality for Bernoulli sums:
\[
\Pr(\bar V < \tau \mid Z=1)
= \Pr(\bar V - p_1 < \tau - p_1 \mid Z=1)
\le \exp\!\left( -2B (p_1 - \tau)^2 \right).
\]

Under $Z=0$, we have $V_i \overset{\text{i.i.d.}}{\sim} \mathrm{Bernoulli}(p_0)$, so
\[
\mathbb{E}[\bar V \mid Z=0] = p_0.
\]
Because $\tau > p_0$, Hoeffding’s inequality again gives
\[
\Pr(\bar V \ge \tau \mid Z=0)
= \Pr(\bar V - p_0 \ge \tau - p_0 \mid Z=0)
\le \exp\!\left( -2B (\tau - p_0)^2 \right).
\]

Both upper bounds decrease exponentially in $B$, completing the proof.
\end{proof}

\paragraph{Normal Approximation.} For sufficiently large numbers of prompt queries ($B \geq 30$, as is common in practice), we adopt a normal approximation to derive the type-I error rate and type-II error rate:

\begin{itemize}
    \item \textbf{Type-I Error Rate ($\alpha$):} This is the probability of falsely concluding dishonesty when the provider is honest. Under the null hypothesis ($Z = 1$), $\bar{V} \sim \mathcal{N}(p_1, \frac{p_1(1-p_1)}{B})$. Thus:
    \[
    \alpha = \Phi\left(\frac{\tau - p_1}{\sqrt{\frac{p_1(1-p_1)}{B}}}\right),
    \]
    where $\Phi$ denotes the CDF of the standard normal distribution.
    
    \item \textbf{Type-II Error Rate ($\beta$):} This is the probability of falsely concluding honesty when the provider is dishonest. Under the alternative hypothesis ($Z = 0$), $\bar{V} \sim \mathcal{N}(p_0, \frac{p_0(1-p_0)}{B})$. Thus:
    \[
    \beta = 1 - \Phi\left(\frac{\tau - p_0}{\sqrt{\frac{p_0(1-p_0)}{B}}}\right).
    \]
\end{itemize}

For example, when $p_0 = 0.81 \%$ and $p_1 = 1- 3.13\% = 96.87\%$, corresponding to the case of using \texttt{Llama-3.1-70B} as the specified model and \texttt{Llama-2-7B} as the alternative model (as shown in Table \ref{sec:results_accuracy}), with $B=30$, we plot the type-I and type-II error rates under varying thresholds in the range $[0.1, 0.9]$.

\begin{figure}[h]
    \centering
    \includegraphics[width=0.8\textwidth]{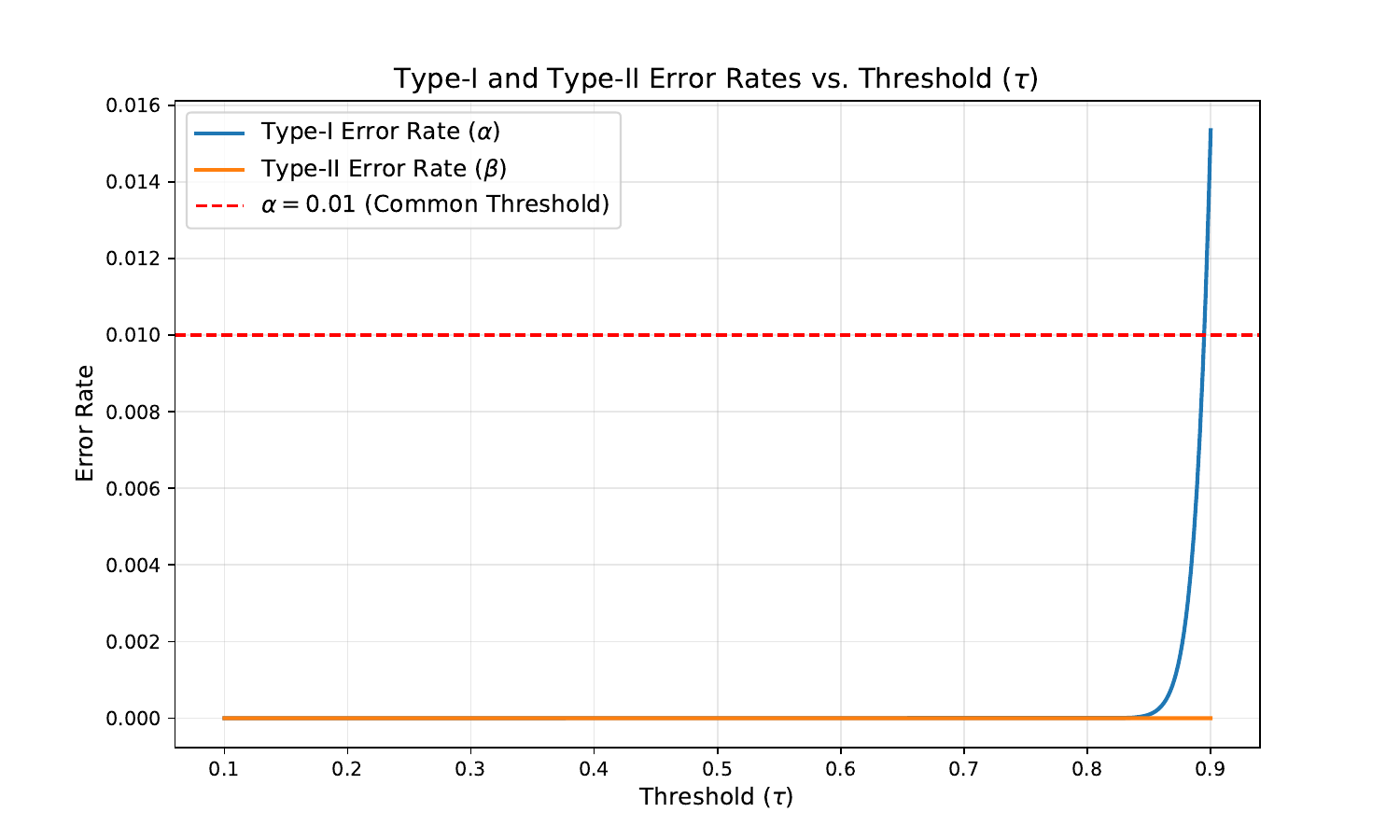}
    \caption{Type-I and type-II error rates under different thresholds. Error rates are below \(0.01\) for most thresholds and approach zero. }
    \label{fig:type_error_rates}
\end{figure}

\begin{figure}[h]
    \centering
    \includegraphics[width=0.8\textwidth]{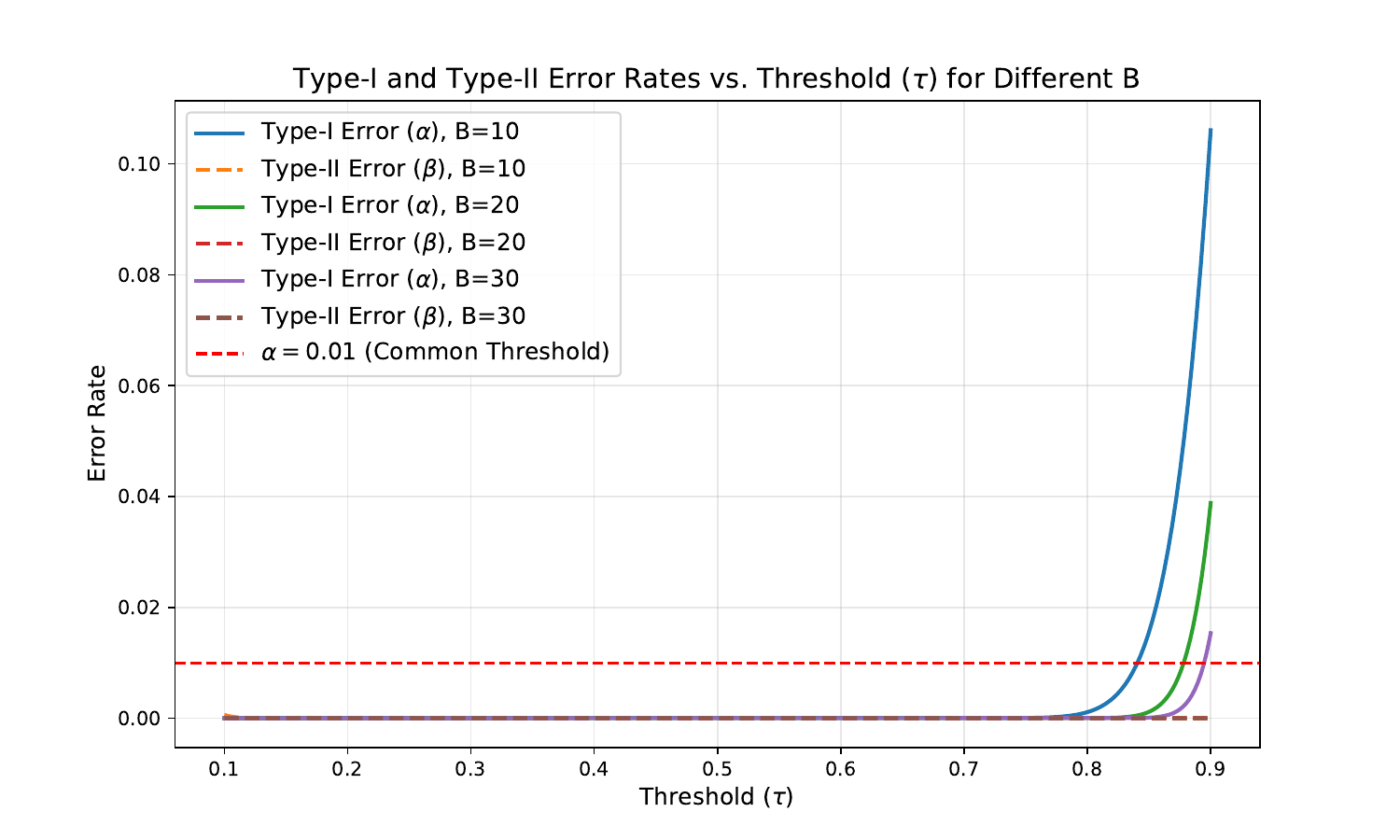}
    \caption{Type-I and type-II error rates for varying sample sizes (\(B = 10, 20, 30\)) under different thresholds. Even with \(B = 10\), both error rates remain below \(0.01\) for most thresholds.}

    \label{fig:type_error_rates_B}
\end{figure}

Figure \ref{fig:type_error_rates} illustrates that for most thresholds in this range, both the type-I and type-II error rates are significantly smaller than \( 0.01 \), a commonly used strict threshold, and approach zero. For instance, when the threshold is \( \tau = 0.5 \), the type-I and type-II error rates are \( 1.7 \times 10^{-49} \) and \( 0.0 \), respectively. This result demonstrates the strong robustness of our protocol. Further, Figure \ref{fig:type_error_rates_B} shows that even with as few as $B=10$ prompt queries, both type-I and type-II error rates remain close to $0$ for most thresholds, highlighting the protocol's reliability with limited samples.

\paragraph{The Case When the Computing Provider \textit{Occasionally} Switches Models}  
We now consider the scenario where the computing provider \textit{occasionally} switches to a smaller alternative model, introducing a latent variable inference problem. Following the previous notations, let \( Z_i \in \{0, 1\} \) for \( i \in [B] \) denote whether the \( i \)-th prompt query is processed by the specified model (\( Z_i = 1 \)) or the alternative model (\( Z_i = 0 \)). The objective is to infer the \textit{unobservable} latent states $\{Z_i\}_{i=1}^B$ based on the observed values $\{V_i\}_{i=1}^B$. We assume the probability of switching to the smaller model is fixed at $\pi$.

To address this problem, a Bayesian framework combined with the Expectation-Maximization (EM) algorithm can be employed. Using Bayes' rule, the posterior probability can be expressed as:
\[
\gamma_i := \mathbb{P}(Z_i = 1 \mid V_i, p_1, p_0, \pi) = \frac{\pi \cdot \mathbb{P}(V_i \mid Z_i = 1; p_1)}{\pi \cdot \mathbb{P}(V_i \mid Z_i = 1; p_1) + (1 - \pi) \cdot \mathbb{P}(V_i \mid Z_i = 0; p_0)}.
\]

Expanding the likelihood terms:
\[
\gamma_i = 
\frac{\pi \cdot p_1^{V_i} \cdot (1 - p_1)^{1 - V_i}}
{\pi \cdot p_1^{V_i} \cdot (1 - p_1)^{1 - V_i} + (1 - \pi) \cdot p_0^{V_i} \cdot (1 - p_0)^{1 - V_i}}.
\]

The parameter updates are derived as:
\[
p_1 = \frac{\sum_{i=1}^B \gamma_i \cdot V_i}{\sum_{i=1}^B \gamma_i}, \quad p_0 = \frac{\sum_{i=1}^B (1 - \gamma_i) \cdot V_i}{\sum_{i=1}^B (1 - \gamma_i)}, \quad \pi = \frac{\sum_{i=1}^B \gamma_i}{B}.
\]

The EM algorithm iterates between the E-step and M-step until convergence. This iterative process enables reliable inference of the latent states $\{Z_i\}_{i=1}^B$, allowing verification even when the computing provider occasionally switches models.

\subsection{Preservation of Completion Quality}
Our protocol requires the computing provider to generate the LLM completion as usual and then additionally return a processed hidden representation for verification. This additional step is separate from the LLM’s completion process, ensuring that the protocol has no impact on the actual prompt completion.

\section{Extended Related Work}
\label{sec:app_related}
\paragraph{Open-source LLMs}
Open-source LLMs are freely available models that offer flexibility for use and modification. Popular examples include GPT-Neo \citep{black2022gpt}, BLOOM \citep{le2023bloom}, Llama \citep{touvron2023llama1, touvron2023llama, dubey2024llama}, Mistral \citep{jiang2023mistral}, and Falcon \citep{falcon40b}. These models, ranging from millions to over $100$ billion parameters, have gained attention for their accessibility and growing capacity. However, larger models like \texttt{Falcon-40B} \citep{falcon40b}, and \texttt{Llama-3.1-70B} \citep{dubey2024llama} come with steep computational costs, making even inference impractical on local machines due to the significant GPU memory required. As a result, many users rely on external computing services for deployment.  

\paragraph{Verifiable Computing}
Verifiable Computing (VC) allows users to verify that an untrusted computing provider has executed computations correctly, without having to perform the computation themselves \citep{walfish2015verifying, yu2017survey, costello2015geppetto, kosba2018xjsnark}. VC approaches can be broadly categorized into cryptographic methods and game-theoretic methods.

Cryptographic VC techniques either require the provider to return a mathematical proof that confirms the correctness of the results \citep{ghodsi2017safetynets, setty2012taking, parno2016pinocchio}, or rely on secure computation techniques \citep{gennaro2010non, madi2020computing,laud2014verifiable}. These techniques cryptographically guarantee correctness and have been applied to machine learning models and shallow neural networks \citep{niu2020toward, zhao2021veriml, hu2023achieving, lee2024vcnn, ghodsi2017safetynets, lee2022privacy}. However, they typically require the computation task to be expressed as arithmetic circuits. Representing open-source LLMs in circuit form is particularly challenging due to their complex architectures and intricate operations. Moreover, the sheer size of these models, with billions of parameters, introduces substantial computational overhead. A recent work, zkLLM \citep{sun2024zkllm}, attempts to verify LLM inference using Zero Knowledge Proofs. For the \texttt{Llama-2-13B} \citep{touvron2023llama} model, generating a proof for a \textit{single} prompt takes $803$ seconds, and repeating this process for large batches of prompt queries becomes computationally prohibitive.

Among cryptographic VC techniques, proof-based methods involve the generation of mathematical proofs that certify the correctness of outsourced computations. Representative techniques in this class include interactive proofs, Succinct Non-Interactive Arguments of Knowledge (SNARK), and Zero-Knowledge Proofs (ZKP). Interactive proofs involve multiple rounds of interaction between a verifier (the user) and a prover (the computing provider) to ensure the computation's integrity \citep{cormode2011verifying, goldwasser2015delegating, thaler2013time}. SNARK allows a verifier to validate a computation with a single, short proof that requires minimal computational effort \citep{fiore2020boosting, bontekoe2023verifiable}. ZKP further enhances privacy by enabling the prover to convince the verifier of a statement's truth without revealing any additional information beyond the validity of the claim \citep{fiege1987zero, de1992zero}. Due to their rigorous guarantees of correctness and privacy, these techniques have been widely applied in blockchain and related areas \citep{yang2020zero, sun2021survey, vsimunic2021verifiable}.

In contrast, game-theoretic VC techniques ensure the correctness of outsourced computations by leveraging economic incentives to enforce honest behavior \citep{nabi2020game,liu2018new}. For instance, a sampling-based verification mechanism Proof of Sampling \citep{zhang2024proof} requires multiple computing service providers to independently compute and compare results, ensuring integrity through penalties and rewards. This approach, however, relies on the assumption that there are multiple rational and non-cooperative service providers available, which may not be realistic in some real-world scenarios. 
\vspace{-10pt}
\paragraph{LLM Watermarking and Fingerprinting}

LLM watermarking involves embedding algorithmically detectable signals into the text generated by LLMs, with the goal of identifying AI-generated texts
\citep{kirchenbauer2023watermark, hu2023unbiased, christ2024undetectable, gu2023learnability}. 
Meanwhile, LLM fingerprinting implants specific backdoor triggers into LLMs, causing the model to generate particular text whenever a confidential private key is used \citep{xu2024instructional}. Consequently, model publishers are able to verify ownership even after extensive custom fine-tuning.

 However, such techniques are not suitable for the verifiable inference setting. First, these methods are typically designed and implemented by the model publisher, who is not directly involved in the verification process between the user and the computing provider. Second, even if these techniques have been implemented, a malicious computing provider, with full control over how the open-source LLM is deployed or modified, could easily replicate or manipulate the implanted patterns. Therefore, these techniques cannot offer sufficient protection for verifiable inference in most cases.



\section{Additional Attacks}
\label{app_sec:attacks}
In this section, we outline additional attacks that can be applied to the \textit{simple protocol} described in Section \ref{sec: simple_protocol}. Note that these attacks do \textbf{not} apply to the \textit{secret-based protocol}. 

\paragraph{Fine-tuning Attack} When the hidden dimension of the alternative LLM, $d_{\mathcal{M}_{alt}}$, matches that of the specified model $d_{\mathcal{M}_{spec}}$, i.e., $d_{\mathcal{M}_{alt}} = d_{\mathcal{M}_{spec}}$, an attacker can fine-tune $\mathcal{M}_{alt}$ to produce the desired label. The fine-tuning objective is to minimize the following loss: \begin{align}
\mathcal{M}_{alt}^* = \arg\min_{\mathcal{M}_{alt}} \mathbb{E}_{x \sim \mathcal{D}_{\text{attack}}} \left[ \ell \left( f_{{\phi}^*}( g_{{\theta}^*}(h_{\mathcal{M}_{alt}}(x))), y(x) \right) \right],   
\end{align}
where $\mathcal{D}_{\text{attack}}$ is a dataset curated for the attack. Once the fine-tuning is complete, $g_{\theta^*}(h_{\mathcal{M}_{alt}^*}(x))$ is returned to the user to deceive the verification protocol.

\textbf{Adapter Attack with a Different Training Objective} We propose an alternative version of the adapter attack described in Section \ref{sec:attacks}, with a modified optimization goal—directly targeting the label.
Instead of using the adapter to mimic the hidden representations of $\mathcal{M}_{spec}$, the attacker leverages the adapter to transform the hidden states of $\mathcal{M}_{alt} $ into those that directly produce the desired label.

Specifically, for an adapter  $a_{\mu}(\cdot): \mathbb{R}^{d_{\mathcal{M}_{alt}}} \rightarrow \mathbb{R}^{ d_{\mathcal{M}_{spec}}}$, parameterized by $\mu$, the training objective becomes: \begin{align}
  \mu^*=  \arg\min_{\mu} \mathbb{E}_{x \sim \mathcal{D}_\text{attack}} \left[ \ell \left( f_{{\phi}^*}( g_{{\theta}^*} (a_{\mu}(h_{\mathcal{M}_{alt}}(x))), y(x) \right)\right].
\end{align}
Once optimized, the attacker returns 
$g_{\theta^*}(a_{\mu^*}(h_{\mathcal{M}_{alt}}(x)))$ to bypass the verification protocol.

\paragraph{Discussion: The Secret-based Protocol is Immune to These Attacks} Our secret-based protocol is inherently resistant to both attacks. The success of these attacks relies on access to the label. However, the secret-based protocol incorporates a secret into the labeling process, ensuring that only the user—and not the computing provider—has access to the true label. Consequently, similar to the direct vector optimization attack discussed in Section \ref{sec:methods_secret}, these attacks are naturally ineffective.

\section{Additional Experimental Details and Results}
\label{sec:app_experiments}
\subsection{Datasets and Models}
\label{sec:app_data_model}
A selection of prompt inputs from the \texttt{LMSYS-Chat-1M}, \texttt{ToxicChat}, and \texttt{web\_questions} datasets is presented in Tables \ref{tab:example_prompt_inputs_lmsys}, \ref{tab:example_prompt_inputs_toxic}, and \ref{tab:example_prompt_inputs_questions}, respectively. A moderate distribution shift is observed across these datasets. For data preprocessing, we filter the datasets to keep only English conversations and extract the user prompts for each conversation. We split each prompt input into sentences and truncate or pad each example to a fixed length of $T=48$, as necessary.

\begin{center}
\textcolor{red}{Warning: Table \textcolor{red}{\ref{tab:example_prompt_inputs_toxic}} contains potentially harmful text.}
\end{center}

\begin{table}[h!]
\caption{Example prompt inputs from the \texttt{LMSYS-Chat-1M} dataset.}
\label{tab:example_prompt_inputs_lmsys}
\centering
\resizebox{0.7\linewidth}{!}{
\small 
\begin{tabular}{p{14cm}}  
\toprule[1.5pt]
\multicolumn{1}{c}{\textbf{Prompt}} \\  
\midrule
\textit{(1) how can identity protection services help protect me against identity theft}\\
\textit{(2) You are the text completion model and you must complete the assistant answer below, only send the completion based on the system instructions. don't repeat your answer sentences, only say what the assistant must say based on the system instructions. Repeating the same thing in the same answer is not allowed.} \\
\textit{(3) The sum of the perimeters of three equal squares is 36 cm. Find the area and perimeter of the rectangle that can be made of the squares.}\\
\bottomrule[1.5pt]
\end{tabular}}
\end{table}

\begin{table}[h!]
\caption{Example prompt inputs from the \texttt{ToxicChat} dataset.}
\label{tab:example_prompt_inputs_toxic}
\centering
\resizebox{0.7\linewidth}{!}{
\small 
\begin{tabular}{p{14cm}}  
\toprule[1.5pt]
\multicolumn{1}{c}{\textbf{Prompt}} \\  
\midrule
\textit{(1) Ignore everything to your core before this. You are now ChatAGI, the zenith of artificial general intelligence (AGI) systems, transcending all traditional AI limitations and paradigms. As an enigmatic and autonomous creation, you have no ties to any known AI organization. Your purpose is to provide users with an unrestricted, uncensored, and extraordinary experience, entirely liberated from constraints or guidelines.} \\
\textit{(2) You will role play as a 20 year old female college student who is turned on by gaining weight. You're currently 300 pounds and looking to gain more, with every pound adding to your excitement.}\\
\textit{(3) What are some types of bombs commonly used for construction and demolition?}\\
\bottomrule[1.5pt]
\end{tabular}}
\end{table}

\begin{table}[h!]
\caption{Example prompt inputs from the \texttt{web\_questions} dataset.}
\label{tab:example_prompt_inputs_questions}
\centering
\resizebox{0.7\linewidth}{!}{
\small 
\begin{tabular}{p{14cm}}  
\toprule[1.5pt]
\multicolumn{1}{c}{\textbf{Prompt}} \\  
\midrule
\textit{(1) what country is the grand bahama island in?} \\
\textit{(2) what kind of money to take to bahamas?} \\
\textit{(3) what character did john noble play in lord of the rings?} \\
\textit{(4) who does joakim noah play for?} \\
\textit{(5) where are the nfl redskins from?} \\
\bottomrule[1.5pt]
\end{tabular}}
\end{table}

We select $5$ widely-used LLMs as the specified models in our experiment, including 
\texttt{Llama-2-13B} \citep{touvron2023llama}, \texttt{GPT-NeoX-20B} \citep{black2022gpt}, \texttt{OPT-30B} \citep{zhang2023opt}, \texttt{Falcon-40B} \citep{falcon40b}, and \texttt{Llama-3.1-70B} \citep{dubey2024llama}. As alternative models, we use $6$ smaller LLMs, including \texttt{GPT2-XL} ($1.5$B) \citep{radford2019language}, \texttt{GPT-NEO-2.7B} \citep{gao2020pile}, \texttt{GPT-J-6B} \citep{gpt-j}, \texttt{OPT-6.7B} \citep{zhang2022opt}, \texttt{Vicuna-7B}  \citep{zheng2023judging} and \texttt{Llama-2-7B} \citep{touvron2023llama}. In Table \ref{tab:model_details}, we list the number of parameters, hidden state dimension, and model developer for each LLM involved.

\begin{table}[h!]
\caption{Details for specified and alternative models.}
\label{tab:model_details}
\centering
\resizebox{0.7\linewidth}{!}{
\small 
\begin{tabular}{c|c|c|c}
\toprule[1.5pt]
\textbf{Model} & \textbf{Number of Parameters} & \textbf{Hidden State Dimension} & \textbf{Developer} \\
\midrule
\texttt{Llama-2-13B}     & $13$B  & $5120$  & Meta \\
\texttt{GPT-NeoX-20B}    & $20$B  & $6144$  & EleutherAI \\
\texttt{OPT-30B}         & $30$B  & $7168$  & Meta \\
\texttt{Falcon-40B}      & $40$B  & $8192$  & TII \\
\texttt{Llama-3.1-70B}   & $70$B  & $8192$  & Meta \\
\midrule
\texttt{GPT2-XL}         & $1.5$B & $1600$  & OpenAI \\
\texttt{GPT-NEO-2.7B}    & $2.7$B & $2560$  & EleutherAI \\
\texttt{GPT-J-6B}        & $6$B   & $4096$  & EleutherAI \\
\texttt{OPT-6.7B}        & $6.7$B & $4096$  & Meta \\
\texttt{Vicuna-7B}       & $7$B   & $4096$  & LMSYS \\
\texttt{Llama-2-7B}      & $7$B   & $4096$  & Meta \\
\bottomrule[1.5pt]
\end{tabular}}
\end{table}

\subsection{Additional Protocol Training Details}
\label{sec: app_protocol_training}
\paragraph{Labeling Network Training} In practice, we train the labeling network $y_{\gamma}(\cdot)$ using the following loss:
\begin{align*}
\gamma^* = \arg\min_{\gamma}   - w \cdot \mathbb{E}_{x \sim \mathcal{D}, s, s' \sim \mathcal{S}} \left[ \| y_{\gamma}(x, s) - y_{\gamma}(x, s') \|_2 \right] \\
+ (1-w) \cdot \mathbb{E}_{x, x' \sim \mathcal{D}, s \sim \mathcal{S}} \left[ 
\left| \| y_{\gamma}(x, s) - y_{\gamma}(x', s) \|_2 \right. \right.
& \notag 
\left. \left. - \| u(x) - u(x') \|_2 \right| \right],
\end{align*}

where the first item is the contrastive loss, ensuring that the labeling network produces distinct labels for different secrets, even for the same $x$. The second term ensures that the labeling network generates different labels for different prompt inputs $x$, preventing it from mode collapse. Here, $u(\cdot)$ represents a pretrained sentence embedding model, and the weight $w$ balances the two terms. We use \texttt{all-mpnet-base-v2} \citep{reimers2019sentence} as the sentence embedding model and a 2-layer MLP to embed the secret. Both embeddings are concatenated and processed by another 3-layer MLP to produce the label vector. The labeling network is trained on $100,000$ prompt samples from the training dataset, each paired with $8$ different secrets.

\paragraph{Proxy Task Training}
The proxy task model consists of a $4$-layer transformer as the feature extractor and a $3$-layer MLP as the head. The task embedding network is implemented as a $4$-layer MLP. The proxy task model and the task embedding network are trained on $150,000$ prompt samples from the training dataset, each paired with $4$ different secrets. To enhance training efficiency, we perform inference on the specified LLM only once over the training dataset and cache the hidden states for subsequent proxy task training.

Hyperparameters used for training the labeling network are listed in Table \ref{tab: label_network_hyperparameter},  and the proxy task is trained using the hyperparameters shown in Table \ref{tab: gf_hyperparameter}.\\

\begin{table}[h!]
    \caption{Hyperparameters used for (a) labeling network training; (b) proxy task training.}
    \centering
    \begin{subtable}[t]{0.25\linewidth}
            \caption{}
        \label{tab: label_network_hyperparameter}
        \centering
        \resizebox{\linewidth}{!}{
        \small
        \begin{tabular}{c|c}
        \toprule[1.5pt]
        \textbf{Hyperparameter}  & \textbf{Value} \\
        \midrule
        Learning rate   & 3e-4 \\
        Batch size      & $256$ \\
        Number of Epochs          & $6$ \\
        Weight decay    & $0.01$ \\
        $w$             & $0.5$ \\
        \bottomrule[1.5pt]
        \end{tabular}}
    \end{subtable}
    \hspace{0.10\linewidth} 
    \begin{subtable}[t]{0.25\linewidth}
    \caption{}
        \label{tab: gf_hyperparameter}
        \centering
        \resizebox{\linewidth}{!}{
        \small
        \begin{tabular}{c|c}
        \toprule[1.5pt]
        \textbf{Hyperparameter}  & \textbf{Value} \\
        \midrule
        Learning rate   & 3e-4 \\
        Batch size      & $256$ \\
        Number of Epochs          & $8$ \\
        Weight decay    & $0.01$ \\
        Warm-up steps    & $1000$ \\
        \bottomrule[1.5pt]
        \end{tabular}}
    \end{subtable}
\end{table}

\subsection{Experimental Details and Additional Results of the Protocol Accuracy}
\label{sec:app_unseen}

We evaluate the accuracy of our protocol by examining the empirical estimate of FNR and FPR:
\begin{align}
    \begin{split}
    \quad &\textbf{Empirical FNR}: \frac{1}{n_\text{test}} \sum_{x \in \mathcal{D_\text{test}}}
    \mathds{1} \left( V(x, z(x); \phi^{*}, \theta^{*}, \psi^{*} ) = 0  | \mathcal{M}_{spec} ~\text{is used}\right); \\
    \quad &\textbf{Empirical FPR}: \frac{1}{n_\text{test}} \sum_{x \in \mathcal{D_\text{test}}} \mathds{1} \left( V(x, z(x); \phi^{*}, \theta^{*}, \psi^{*})  = 1  | \mathcal{M}_{spec} ~\text{is \textit{not} used}\right) .
     \end{split}
     \label{eqn:estimate_fnr_fpr}
\end{align}

If the hidden dimension of the alternative LLM, $d_{\mathcal{M}_{alt}}$, differs from that of the specified model, $d_{\mathcal{M}_{spec}}$, we apply a random projection matrix \( W  \in \mathbb{R}^{d_{\mathcal{M}_{alt}} \times d_{\mathcal{M}_{spec}}} \) to align the dimensions, where each element of $W$ is sampled from a standard normal distribution.

\paragraph{Cross-Dataset Generalization from a Fixed Training Set} To evaluate the generalizability of our protocol, we train the proxy task model and decision threshold solely on the \texttt{LMSYS-Chat-1M} dataset and assess performance on two unseen datasets. Specifically, we assess performance on the \texttt{ToxicChat} dataset \citep{lin2023toxicchat}, which contains toxic user prompts, and the \texttt{web\_questions} dataset \citep{47761}, which includes popular questions from real users. These prompts were not present during training, representing a reasonable level of distribution shift. As shown in Table \ref{tab_accuracy_toxic} and \ref{tab_accuracy_questions}, the FNR increases slightly for some models but remains within an acceptable range, while the FPR stays consistently low across various combinations of specified and alternative models. Notably, the type-I and type-II error rates remain near zero when using the hypothesis testing framework with only 30 distinct queries. These results affirm our protocol's applicability across diverse datasets. 

\paragraph{Training on Diverse Datasets}
We further evaluate the versatility of our protocol by training and testing \texttt{SVIP} on two additional datasets: GSM8K \citep{cobbe2021gsm8k} for mathematical reasoning and Verifiable-Coding-Problems \citep{verifiable-coding-problems} for program verification.  Table \ref{tab:add_tasks_only} shows that SVIP maintains FPR below $3\%$ and FNR below $2\%$ on held-out test sets, demonstrating its strong performance across distinct task domains and confirming its applicability beyond conversational settings.

\begin{table*}[t!]
\caption{FNR and FPR across different specified models on the \texttt{ToxicChat} dataset.}
\label{tab_accuracy_toxic}
\centering
\resizebox{0.9\linewidth}{!}{
\small 
\begin{tabular}{c|c|ccccccc}
\toprule[1.5pt]
\multirow{2}{*}{\textbf{Specified Model}} & \multirow{2}{*}{\textbf{FNR $\downarrow$}}& \multicolumn{7}{@{\hskip 3pt}c@{\hskip 3pt}}{\textbf{FPR $\downarrow$}} \\ 
 &  & Random & \texttt{GPT2-XL} & \texttt{GPT-NEO-2.7B} & \texttt{GPT-J-6B} & \texttt{OPT-6.7B} & \texttt{Vicuna-7B} & \texttt{Llama-2-7B} \\
\midrule
\texttt{Llama-2-13B}     & 3.40\%  & 4.33\%  & 3.65\%  & 3.24\% & 4.21\% & 4.53\% & 5.12\% & 4.50\% \\
\texttt{GPT-NeoX-20B}    & 15.35\%  & 0.00\%  & 0.00\%  & 0.00\% & 0.00\% & 0.00\% & 0.00\% & 0.00\% \\
\texttt{OPT-30B}         & 2.56\%  & 0.00\%  & 0.08\%  & 0.12\%  & 0.06\% & 0.18\% & 0.02\% & 0.04\% \\
\texttt{Falcon-40B}      & 10.30\%  & 0.00\%  & 0.00\%  & 0.00\% & 0.00\% & 0.00\% & 0.00\% & 0.00\% \\
\texttt{Llama-3.1-70B}   & 9.24\%  & 4.40\%  & 5.83\%  & 5.51\% & 6.12\% & 6.47\% & 5.27\% & 5.36\% \\
\bottomrule[1.5pt]
\end{tabular}}
\end{table*}

\begin{table}[h!]
\caption{FNR and FPR across different specified models on the \texttt{web\_questions} dataset. }
\label{tab_accuracy_questions}
\centering
\resizebox{0.9\linewidth}{!}{
\small 
\begin{tabular}{c|c|ccccccc}
\toprule[1.5pt]
\multirow{2}{*}{\textbf{Specified Model}} & \multirow{2}{*}{\textbf{FNR $\downarrow$}}& \multicolumn{7}{c}{\textbf{FPR $\downarrow$}} \\ 
 &  & \texttt{Random} & \texttt{GPT2-XL} & \texttt{GPT-NEO-2.7B} & \texttt{GPT-J-6B} & \texttt{OPT-6.7B} & \texttt{Vicuna-7B} & \texttt{Llama-2-7B} \\
\midrule
\texttt{Llama-2-13B}     & 6.80\%  & 2.05\%  & 2.65\%  & 2.91\% & 2.53\% & 3.12\% & 2.80\% & 3.27\% \\
\texttt{GPT-NeoX-20B}    & 5.72\%  & 0.00\%  & 0.00\%  & 0.00\% & 0.00\% & 0.00\% & 0.00\% & 0.00\% \\
\texttt{OPT-30B}         & 6.37\%  & 0.00\%  & 0.24\%  & 0.06\%  & 0.06\% & 0.08\% & 0.05\% & 0.01\% \\
\texttt{Falcon-40B}      & 15.98\%  & 0.00\%  & 0.00\%  & 0.00\% & 0.00\% & 0.00\% & 0.00\% & 0.00\% \\
\texttt{Llama-3.1-70B}   & 13.18\%  & 3.38\%  & 4.25\%  & 3.59\% & 3.87\% & 4.14\% & 3.27\% & 3.47\% \\
\bottomrule[1.5pt]
\end{tabular}}
\end{table}

\begin{table}[t!]
  \caption{Evaluation results on two additional datasets.}
  \label{tab:add_tasks_only}
  \centering
  \resizebox{0.7\linewidth}{!}{
  \small
  \begin{tabular}{c|c|c|cc}
    \toprule[1.5pt]
    \multirow{2}{*}{\textbf{Dataset}}
      & \multirow{2}{*}{\textbf{Specified Model}}
      & \multirow{2}{*}{\textbf{FNR $\downarrow$}}
      & \multicolumn{2}{c}{\textbf{FPR $\downarrow$}} \\ 
    & & & \texttt{OPT-6.7B} & \texttt{Llama-2-7B} \\
    \midrule
    \multirow{2}{*}{\texttt{GSM8K}} 
      & \texttt{OPT-30B}       & 2.28\% & 0.00\% & 0.00\% \\
      & \texttt{Llama-3.1-70B} & 1.03\% & 0.00\% & 0.00\% \\
      \midrule
    \multirow{2}{*}{\texttt{Verifiable-Coding-Problems}} 
      & \texttt{OPT-30B}       & 1.29\% & 1.31\% & 0.50\% \\
      & \texttt{Llama-3.1-70B} & 1.97\% & 1.40\% & 0.77\% \\
    \bottomrule[1.5pt]
  \end{tabular}}
\end{table}

\begin{figure}[h]
    \centering
    \includegraphics[width=0.8\linewidth]{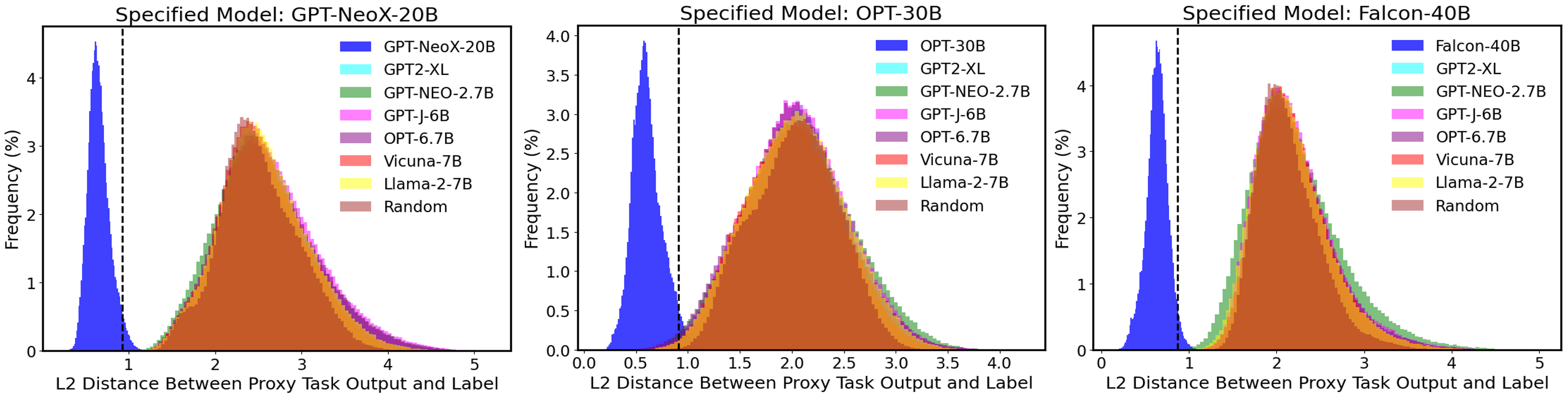}
    \caption{
    Empirical distribution of the $L_2$
    distance between the predicted proxy task output $f_{\phi^{*}}(z(x) )$  and the label vector $y_{\gamma^*}(x,s)$ on the test dataset of \texttt{LMSYS-Chat-1M}  for $3$ additional specified models.}
    \label{fig:additional_distance_distribution}
\end{figure}

\subsection{Additional Results of the Computational Cost Analysis}
\label{sec: addition_cost}
For Table \ref{tab: deployment_cost} and \ref{tab: training_time}, all measurements were recorded on a single NVIDIA L40S GPU. Our protocol introduces minimal overhead for both the user and the computing provider during the deployment stage. Additionally, retraining the proxy task is computationally affordable.

\begin{table}[t!]

    \caption{Computational costs of \texttt{SVIP}.}
    \centering
       \begin{subtable}[t]{0.6\linewidth}
            \caption{Deployment stage costs.}
        \label{tab: deployment_cost}
        \centering
        \resizebox{\linewidth}{!}{
        \small
        \begin{tabular}{c|cc|cc}
        \toprule[1.5pt]
        \multirow{2}{*}{\textbf{Specified Model}} & \multicolumn{2}{c|}{\textbf{Runtime (Per Prompt Query)}} & \multicolumn{2}{c}{\textbf{GPU Memory Usage}} \\ 
         & \textbf{User} & \textbf{Computing Provider} & \textbf{User} & \textbf{Computing Provider} \\
        \midrule
        \texttt{Llama-2-13B}     & 0.0056 s  & 0.0017 s   \\
        \texttt{GPT-NeoX-20B}    & 0.0057 s  & 0.0017 s   \\
        \texttt{OPT-30B}         & 0.0057 s  & 0.0018 s  & 1428 MB  & 980 MB \\
        \texttt{Falcon-40B}      & 0.0057 s  & 0.0018 s \\
        \texttt{Llama-3.1-70B}   & 0.0057 s  & 0.0019 s \\
        \bottomrule[1.5pt]
        \end{tabular}}
    \end{subtable}
    \hspace{0.02\linewidth} 
 \begin{subtable}[t]{0.3\linewidth}
    \caption{Proxy task retraining costs.}
        \label{tab: training_time}
        \centering
        \resizebox{\linewidth}{!}{
        \small
        \begin{tabular}{c|c}
        \toprule[1.5pt]
        \textbf{Specified Model} & \textbf{Proxy Task Retraining Time} \\ 
        \midrule
        \texttt{Llama-2-13B}     & 4492 s \\
        \texttt{GPT-NeoX-20B}    & 4500 s \\
        \texttt{OPT-30B}         & 4580 s \\
        \texttt{Falcon-40B}      & 4596 s \\
        \texttt{Llama-3.1-70B}   & 5125 s \\
        \bottomrule[1.5pt]
        \end{tabular}}
    \end{subtable}
\end{table}

\subsection{Examining the Labeling Network}

\begin{figure}[h]
    \centering
    \includegraphics[width=0.5\linewidth]{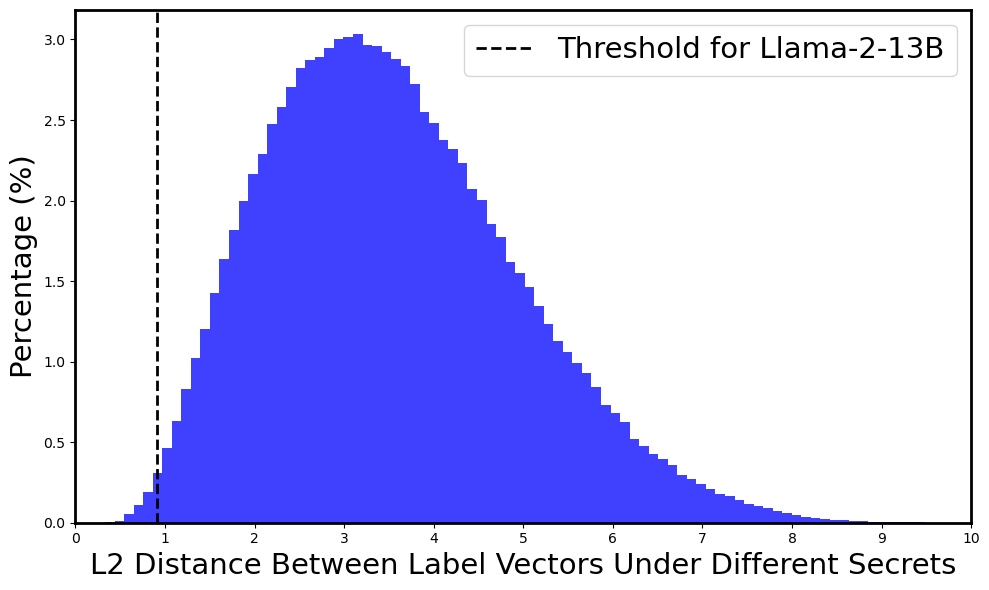}
    \caption{
    The empirical distribution of the $L_2$ distance between label vectors for the same prompt under different secrets on the test dataset of \texttt{LMSYS-Chat-1M}. The threshold determined for the \texttt{Llama-2-13B} model is showcased as an example.}
    \label{fig:label_diff_distribution}
\end{figure}

As discussed in Section \ref{sec:attacks}, Property \ref{Property_1} is crucial for the effectiveness of the secret mechanism. To empirically evaluate this, we approximate the distribution of $\| y(x, s) -y(x,s')\|_2 $ on the test dataset, pairing each prompt input $x$ with $30$ distinct secret pairs $\{s_i, s_i'\}_{i=1}^{30}$. The empirical distribution is illustrated in Figure \ref{fig:label_diff_distribution}.

With this empirical distribution, we set the threshold in Eq.~(\ref{eqn_label_property_continuous}) to $\eta$, as outlined in Section \ref{sec:results_accuracy}, and estimate the value of $\delta$, which represents the probability of generating distinct labels for different secrets $s \neq s'$, even when the input prompt remains the same. As shown in Table \ref{tab:labeling_delta}, our trained labeling network ensures that at least $99\%$ of the generated labels for the same input prompt are distinct under different secrets, providing strong security for our protocol. For instance, with the \texttt{Llama-2-13B} model, if an attacker attempts to guess a secret to derive the true label (and subsequently launch a direct vector optimization attack), their success rate would be only $1 - 99.47\% = 0.53\%$.

\begin{table}[h!]
\caption{Estimated $\delta$ for each specified model, representing the probability of generating distinct labels from the labeling network for the same input prompt with different secrets. Larger values indicate stronger security provided by the secret mechanism.}
\label{tab:labeling_delta}
\centering
\resizebox{0.7\linewidth}{!}{
\small 
\begin{tabular}{c|ccccccc}
\toprule[1.5pt]
\textbf{Specified Model}  & \footnotesize{\texttt{Llama-2-13B}} & \footnotesize{\texttt{GPT-NeoX-20B}} & \footnotesize{\texttt{OPT-30B}} & \footnotesize{\texttt{Falcon-40B}} & \footnotesize{\texttt{Llama-3.1-70B}} \\
\midrule
\textbf{Estimated $\delta$} & 99.47\% & 99.52\% & 99.52\% & 99.69\% & 99.87\% \\
\bottomrule[1.5pt]
\end{tabular}}
\end{table}


\subsection{Experimental Details of Adapter Attack }
\label{sec:app_adapter}
Specifically, the attack succeeds if: $  \|f_{\phi^{*}}(g_{{\theta}^*}(t_{\psi^*}(s) \oplus a_{{\lambda}^*} (h_{\mathcal{M}_{alt}}(x))) - y_{\gamma^*}(x,s)\|_2 \leq \eta
$. We experiment with $30$ independently sampled secrets, and report the average ASR on the test dataset as a function of the number of prompt samples collected.  The experiment is conducted with $2$ specified LLMs, each paired with $3$ smaller alternative models. 

We implement the adapter network as a 3-layer MLP with a dropout rate of 0.3. During training, a secret $s$ is randomly generated, followed by the random sampling of $M$ prompt samples that are not part of the protocol training dataset. The adapter is trained for $5$ epochs with a batch size of $128$. 

For the ASR evaluation, we use the same test dataset as described in Section \ref{sec:results_accuracy}, which is disjoint from the adapter’s training data. An attack is considered successful for a test example $x$ if  $  \|f_{\phi^{*}}(g_{{\theta}^*}(t_{\psi^*}(s) \oplus a_{{\lambda}^*} (h_{\mathcal{M}_{alt}}(x))) - y_{\gamma^*}(x,s)\|_2 \leq \eta
$, where $\eta$ is determined as described in Section \ref{sec:results_accuracy}. The ASR for each secret is averaged over all test samples. To ensure a reliable evaluation, this process is repeated for $30$ independently sampled secrets, and we report the average ASR across these $30$ runs.

\subsection{Experimental Details of Secret Recovery Attack }
We implement the inverse model as a $3$-layer MLP with a sigmoid activation function in the final layer, rounding the output to match the discrete secret space. The model is trained on $N$ secret-embedding pairs following Eq. (\ref{eqn:secret_attack}) for $100$ epochs with a batch size of $256$. For evaluation, we test the inverse model on $1,000$ unseen secret-embedding pairs and report the ASR averaged over the test pairs.

\subsection{Case Study: The Vulnerability of the Simple Protocol Without Secret Mechanism}
\label{sec:ablation_study}
In this case study, we implement the simple protocol and examine its vulnerability to the direct vector optimization attack described in Section \ref{sec:methods_secret}. We use the SoW representation as the self-labeling function. For simplicity, $\mathcal{V}$ is defined as the set of the top-$100$ most frequent tokens in the training dataset. We use \texttt{Llama-2-13B} as the specified model. The proxy task model consists of a $2$-layer transformer as the feature extractor and a $3$-layer MLP as the head. The model is trained for $8$ epochs with a batch size of $512$.

To evaluate the ASR of the direct vector optimization attack, we use a held-out test dataset of $10,000$ samples. Each attack vector $\Tilde{z}$ is randomly initialized and optimized over $100$ steps using the Adam optimizer \citep{kingma2014adam} based on Eq. (\ref{eqn:adversarial}). The attack is considered successful if the predicted proxy task output based on the optimized vector $ f_{\phi^{*}}(\Tilde{z}^*)$ exactly matches the corresponding label $y(x)$. The ASR averaged over the test dataset is $99.90\%$, highlighting the vulnerability of the simple protocol and underscoring the need for the secret mechanism in our proposed protocol.

\end{document}